\documentclass[letterpaper]{article} 

\usepackage[]{aaai2027} 
\usepackage{times} 
\usepackage{helvet} 
\usepackage{courier} 
\usepackage[hyphens]{url} 
\usepackage{graphicx} 
\usepackage{natbib} 
\usepackage{caption} 
\usepackage{amsmath} 
\usepackage{amssymb} 
\usepackage{tikz} 
\usepackage[ruled,vlined]{algorithm2e}
\SetKwInput{KwIn}{Input}
\SetKwInput{KwOut}{Output}

\usepackage{newfloat}
\usepackage{listings}
\DeclareCaptionStyle{ruled}{labelfont=normalfont,labelsep=colon,strut=off} 
\usepackage{xltabular}
\usepackage{pdflscape}
\copyrighttext{Author's version. Accepted for publication at the International
AAAI Conference on Web and Social Media (ICWSM 2027). This is not the version
of record.}

\title{Whose Assessment of Distress? Community Perspectives and LLM Alignment on Well-Being Posts}
\author{
Andrew Aquilina\textsuperscript{\rm 1},
Xiang Lorraine Li\textsuperscript{\rm 1},
Yu-Ru Lin\textsuperscript{\rm 1}
}
\affiliations{
\textsuperscript{\rm 1}School of Computing and Information, University of Pittsburgh\\
\{andrew.aquilina, xianglli, yurulin\}@pitt.edu
}

\usepackage{lipsum}
\usepackage[table]{xcolor}
\usepackage{booktabs}
\usepackage{tabularx}
\usepackage{array}
\usepackage{multirow}
\usepackage{mdframed}
\usepackage{subcaption}

\pgfdeclarepatternformonly{diagHatch}
{\pgfqpoint{0pt}{0pt}}{\pgfqpoint{10pt}{10pt}}{\pgfqpoint{10pt}{10pt}}%
{%
\pgfsetlinewidth{0.25pt}%
\pgfpathmoveto{\pgfqpoint{0pt}{0pt}}%
\pgfpathlineto{\pgfqpoint{10pt}{10pt}}%
\pgfusepath{stroke}%
}

\newcommand{\diagempty}{%
\tikz[baseline=(bb.base)]{
\node[inner sep=0pt, outer sep=0pt,
minimum width=\linewidth,
minimum height=3.0ex,
anchor=base] (bb) {};
\path[pattern=diagHatch, pattern color=black!35]
(bb.south west) rectangle (bb.north east);
}%
}

\usepackage[most]{tcolorbox}
\usepackage{xcolor}
\newcommand{\answerYes}[1]{\textcolor{blue}{#1}}
\newcommand{\answerNo}[1]{\textcolor{teal}{#1}}
\newcommand{\answerNA}[1]{\textcolor{gray}{#1}}

\newtcolorbox{participantquote}{
unbreakable, 
colback=black!3,
colframe=black!55,
boxrule=0.4pt,
borderline west={2.2pt}{0pt}{black!70},
left=6pt,right=6pt,top=4pt,bottom=4pt,
arc=2pt
}

\newtcolorbox{igparticipantquote}{
unbreakable,
colback=green!8,
colframe=green!45!black,
boxrule=0.4pt,
borderline west={2.2pt}{0pt}{green!55!black},
left=6pt,right=6pt,top=4pt,bottom=4pt,
arc=2pt
}

\newtcolorbox{postexcerpt}{
unbreakable,
colback=black!1,
colframe=black!30,
boxrule=0.4pt,
borderline west={1.2pt}{0pt}{black!40},
left=6pt,right=6pt,top=4pt,bottom=4pt,
arc=6pt,
fontupper=\small\ttfamily
}

\newcommand{\Participant}[1]{\begin{participantquote}#1\end{participantquote}}
\newcommand{\IGParticipant}[1]{\begin{igparticipantquote}#1\end{igparticipantquote}}
\newcommand{\Post}[1]{\begin{postexcerpt}#1\end{postexcerpt}}

\newtcolorbox{tracequote}{
unbreakable,
colback=blue!3,
colframe=blue!25!black,
boxrule=0.4pt,
borderline west={2.2pt}{0pt}{blue!40!black},
left=6pt,right=6pt,top=4pt,bottom=4pt,
arc=2pt,
fontupper=\small\itshape
}
\newcommand{\Trace}[1]{\begin{tracequote}{{}}\enspace #1\end{tracequote}}

\usepackage{tikz}
\usetikzlibrary{patterns.meta}

\newcommand{\sevfont}{\fontfamily{phv}\fontseries{b}\selectfont}

\newcommand{\sevbox}[3][]{%
\begingroup
\setlength{\fboxsep}{.18em}%
\tikz[baseline=(B.base)]{%
\node[inner sep=0pt, outer sep=0pt] (B) {%
\colorbox{#2}{%
\makebox[1.45em][c]{%
\sevfont\fontsize{8.6}{8.6}\selectfont
\textcolor{white}{#3}
}%
}%
};
\if\relax\detokenize{#1}\relax\else
\node[overlay, anchor=south east, xshift=.8em, yshift=-.8em]
at (B.south east) {\smash{#1}};
\fi
}%
\endgroup
}
\newcommand{\Nsev}{\sevbox{gray!65}{N}}
\newcommand{\Msev}{\sevbox{yellow!70!black}{M}}
\newcommand{\Mpsev}{\sevbox{red!75}{M+}}

\newcommand{\NsevSS}{\sevbox[\ssmark]{gray!65}{N}}

\newcommand{\MpsevSS}{\sevbox[\ssmark]{red!75}{M+}}

\newcommand{\ssmark}{%
\tikz[baseline=-0.6ex]{
\node[draw=none, fill=cyan!55!blue, circle, inner sep=0.22ex] {%
\textcolor{white}{\sevfont\fontsize{4.2}{4.2}\selectfont S}%
};
}%
}

\begin{document}

\maketitle

\begin{abstract}
Judgments about psychological distress are socially situated: what counts as concerning hinges on community norms around emotional expression, vulnerability, and help-seeking. Yet large language models (LLMs) used for distress detection are typically aligned to a single, undifferentiated standard. How well do these models capture the perspectives of the communities whose language they assess? We address this question through a perspectivist annotation study in which 321 participants provided 9,587 judgments on 1,198 Reddit posts spanning six identity-based communities, yielding community-specific labels. Raters in the contextualized in-group condition show a modest tendency to agree more with their community than uncontextualized out-group raters (OR = 1.18), an effect varying significantly across communities. We then evaluate nine open-weight LLM configurations and four frontier configurations against these labels. Open-weight LLMs systematically over-estimate distress: when communities perceive none-to-mild distress, these models achieve only 31--44\% accuracy, predominantly producing false positives. GPT-5 and Gemini~2.5~Pro show the same none-to-mild inflation even when their full-sample over/under rates are mixed, while Claude Opus~4 is more conservative. This pattern does not simply mirror an outsider reading position: uncontextualized out-group human aggregates were nearly symmetric, with 18\% over-estimation versus 19\% under-estimation. Instead, the models that inflate none-to-mild cases exhibit a distress prior that exceeds both contextualized in-group and uncontextualized out-group human judgments. These findings have implications for equitable AI deployment in mental health contexts, where miscalibrated distress detection may unevenly affect the communities being assessed.
\end{abstract}

\section{Introduction}
\label{intro}

\begin{figure}[!htb]
\centering
\includegraphics[width=0.9\linewidth]{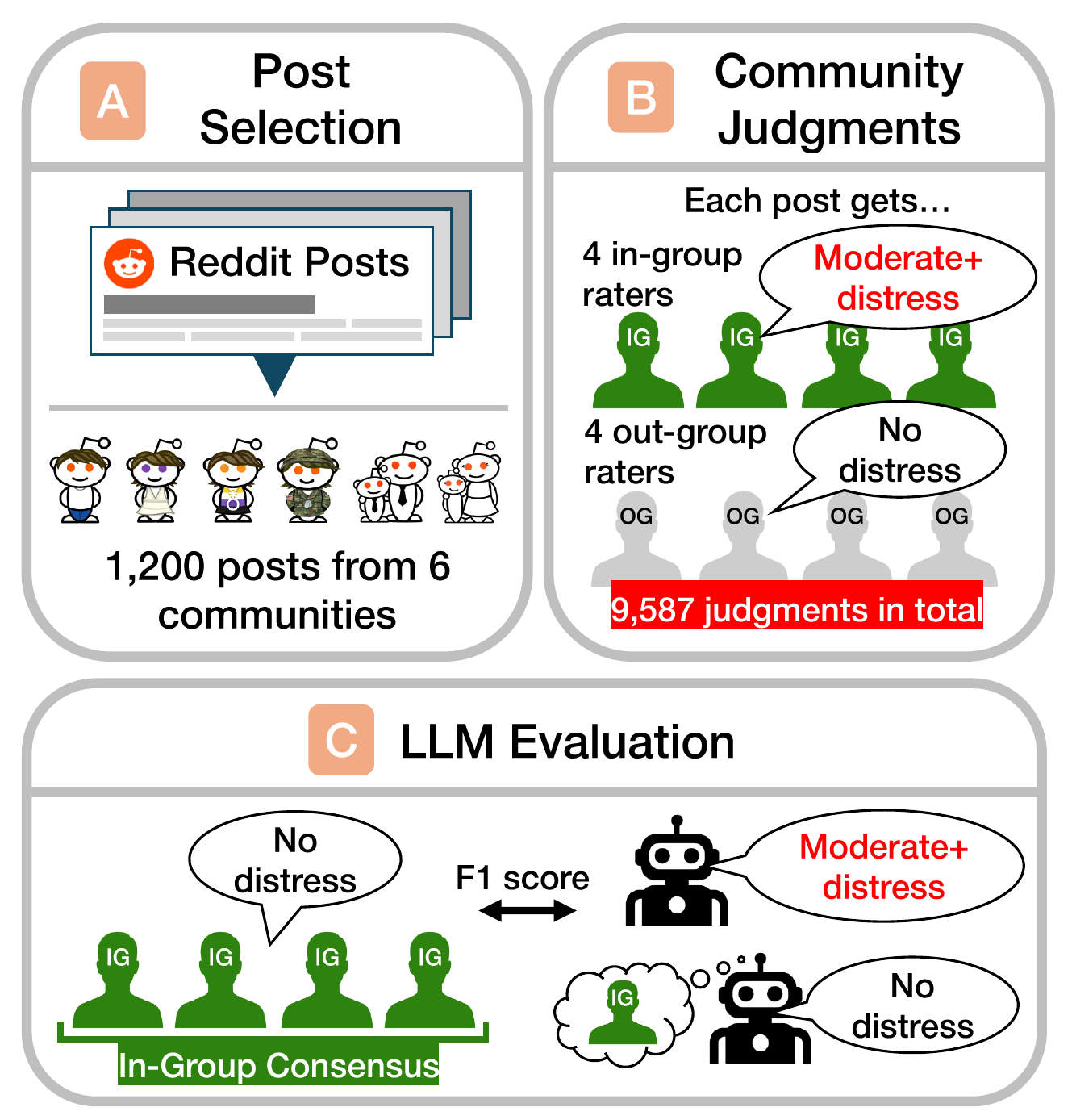}
\caption{Human raters and LLMs read posts sampled from Reddit and judge whether the author is experiencing distress and/or is seeking support. Our paper examines how these judgments vary across communities and how closely LLMs align with community-specific judgments.}
\label{fig:distress-task}
\end{figure}

\textbf{Content Warning: This paper discusses mental health topics that may be distressing to some readers.}
Expressions of psychological distress increasingly appear in the digital traces people leave online. Understanding these expressions has become central to research seeking to offer insight into population mental health trends and support timely interventions. Yet despite growing interest, automated systems for detecting distress still face fundamental challenges: linguistic variability across communities, cultural and contextual nuance, and the risk of reinforcing clinical or demographic biases \cite{rai2025cross,gabriel2024can}. The recent availability of large language models (LLMs) has renewed both excitement and concern. For example, their generative capabilities raise the prospect of adaptive, low-cost mental health support \cite{guo2024large}. However, deploying such systems in sensitive well-being contexts introduces substantial risks, given that judgments about mental health and distress are socially situated and vary across communities  \cite{de2017gender, laws2019differences, treichler2023military}. For these LLM-based tools to be accurate, equitable, and acceptable to users, the interpretations of distress they apply should be grounded in the perspectives of the people whose words are being analyzed. When judgments instead reflect outsider assumptions, models can miss or misread cues and thereby amplify existing inequities.\footnote{Code and data: \url{https://github.com/andaqu/whose-distress}.}

This raises a central concern: \textit{whose judgment of distress is embedded in AI models?} Since distress assessment is inherently interpretive and shaped by community norms, conventional pipelines that optimize toward an ``average'' user's preferences \cite{bakker2022fine} implicitly assume a single standard for what counts as distress. This leads models to miss genuine crises \cite{archiwaranguprok2025simulating} or escalate in alienating ways \cite{ajmani2025seeking}. The recent \textbf{\textit{perspectivist turn}} in NLP offers a promising response, arguing that disagreement among annotators reflects meaningful differences in interpretation rather than noise, and that ``ground truth'' is conditioned on social position, lived experience, and context \cite{fleisig2024perspectivist}.

\textbf{Research focus.} Given that distress and support-seeking judgments are \textbf{\textit{pluralistic}}, shaped by lived experience and community norms rather than universal criteria \cite{mitchell2021well, french2025mental}, we adopt a community-specific evaluation framework to assess whether LLMs calibrate distress severity appropriately or exhibit systematic biases relative to contextualized in-group and uncontextualized out-group human norms.

Specifically, our study examines community judgments of two related but distinct dimensions of mental health expression: \textbf{\textit{distress}} and \textbf{\textit{support-seeking}}. Distress is a psychological response to demands producing an adverse emotional state, often including symptoms such as stress, anxiety, and depression \cite{button2017seeking}. Support-seeking concerns efforts to elicit social resources from others in response to distress \cite{cohen2004social}. To examine how community membership shapes these judgments, we study communities with distinct lived experiences and norms: men and women \cite{chaplin2008gender}, non-binary individuals \cite{darwin2017doing}, veterans \cite{markowitz2023military}, mothers \cite{elliott2020utilization}, and fathers \cite{pedersen2021wanted}. Table~\ref{tab:severity-by-group} illustrates the divergences our analysis surfaces.

Building on this perspectivist framework, we investigate how perceptions of distress and support-seeking vary across communities and how well LLMs capture these judgments. We focus on text-based social media discussion posts, written for community audiences and embedded in shared contexts that support interpreting expressions of distress and support-seeking. Figure~\ref{fig:distress-task} summarizes our study design: we collect \textbf{\textit{in-group}} judgments from raters in a \textit{contextualized} condition, where raters are aware of the post's source community and can draw on shared norms, and \textbf{\textit{out-group}} judgments from raters in an \textit{uncontextualized} condition, where raters assess posts without any community cues. We derive community-specific aggregate labels from these judgments and test how closely LLMs reproduce them. Our research questions are as follows:

\begin{itemize}
\item \textbf{RQ1a:} How do communities differ in their judgments of whether a social media post expresses psychological/emotional distress or seeks support?
\item \textbf{RQ1b:} How does agreement with the communities' aggregate judgment differ between contextualized in-group raters and uncontextualized out-group raters?
\end{itemize}

In RQ1a, we hypothesize that within each author community, labeling distributions will differ between contextualized in-group raters and uncontextualized out-group raters (\textbf{H1a}) \cite{matsumoto1990cultural}. In RQ1b, we expect raters in the contextualized in-group condition to agree with their community’s aggregate label more often than uncontextualized out-group raters judging the same posts (\textbf{H1b}) \cite{elfenbein2002universality}. RQ1 establishes the empirical baseline and alignment targets. Even modest or community-specific differences between these reading conditions would indicate that distress perception is not fully uniform across interpretive contexts, motivating RQ2:

\begin{itemize}
\item \textbf{RQ2a:} To what extent and under what conditions do LLMs align with contextualized in-group judgments on well-being posts?
\item \textbf{RQ2b:} How does identity-based prompting affect model judgments?
\end{itemize}

In RQ2a, we hypothesize that LLMs will vary in how well they align with different groups’ aggregated in-group judgments (\textbf{H2a}), with the greatest misalignment for minority communities (veterans, women, non-binary), following from training-data coverage and preference modeling biases \cite{santurkar2023whose, bakker2022fine}. In RQ2b, we hypothesize that providing group identity and relevant context will improve model alignment across all groups (\textbf{H2b}) \cite{hu2024quantifying}.

To address these RQs, we compile a multi-community dataset of social media posts annotated for distress and support-seeking, and use it to evaluate how well LLMs capture such judgments. Our contributions are as follows:

\begin{enumerate}
\item \textbf{Community-grounded evaluation framework.} We build a multi-community Reddit dataset with a pre-registered\footnote{\url{https://osf.io/3wnyz/overview}} perspectivist annotation study, yielding community-specific aggregate labels against which we evaluate LLMs.
\item \textbf{Reading condition shapes distress judgments.} In-group raters in a contextualized reading condition are modestly more likely than uncontextualized out-group raters to agree with their community's distress aggregate (OR~=~1.18), with significant heterogeneity across communities in both magnitude and, for support-seeking, direction. Through decomposing participants' familiarity with their assigned community, we show that this is not reducible to prior subreddit experience.
\item \textbf{LLMs exhibit an inflated distress prior.} We show that open-weight LLMs systematically over-estimate the severity of distress relative to both in-group and out-group aggregate judgments, a pattern that holds across model scales and families. Among frontier systems, GPT-5 and Gemini~2.5~Pro still over-estimate at the none-to-mild boundary, whereas Claude Opus~4 is conservative and does not show a net upward error. Conditioning models on identity and contextual information reduces over-estimation, but can also induce systematic under-estimation, suggesting that persona-based prompting alone does not resolve the underlying miscalibration.
\end{enumerate}

\section{Related Work}
\label{related-work}

\paragraph{Mental well-being distress detection.} Early work on mental‑health detection commonly drew on social‑media text from Reddit, Twitter, and online support groups. These studies typically paired lexicon-based features with SVM and logistic-regression classifiers \cite{cohan2018smhd, zirikly2019clpsych}. More recent work has shifted to transformer models and instruction-tuned LLMs. For instance, \citet{xu2024mental} fine-tuned open-weight models using social media datasets to predict depression, anxiety, PTSD, and suicidality, out-performing proprietary models such as GPT-3.5 and GPT-4. On the other hand, \citet{settanni2025assessing} evaluated GPT-4o, Claude 3.5 Sonnet, and Gemini 1.5 Pro on triaging Reddit posts by urgency of psychological distress, finding strong correlation with clinician assessments. However, they also found that LLMs tended to over-estimate distress severity, highlighting how their judgments require careful human oversight.

In a recent CLPsych’s shared task \cite{chim2024overview}, researchers operationalized this need for transparency by assessing models’ abilities to justify suicide-risk labels that support a clinician-assigned risk level (low/moderate/high). Analysis showed that performance dropped for low-to-moderate cases due to indirect or subtle linguistic signals that cultural variations in expression may further amplify. Prior work echoes this concern, documenting cross-cultural differences in help-seeking and psychosocial framing on social media across gender and national groups \cite{rai2025cross, de2017gender}. Consequently, recent surveys have emphasized the under-representation of many demographic and cultural groups in this research space \cite{guo2024large, gabriel2024can}, despite evidence that expressions of psychological distress and help-seeking norms vary widely across social groups.

\paragraph{Socio-demographic differences in expressing and perceiving distress.} Psychological distress manifests differently across social and demographic groups. Women tend to express distress more openly through emotions like sadness and worry, whereas men suppress these in favor of anger or stoicism, are less likely to seek help, and are less likely to be diagnosed with depression despite identical symptoms \cite{chaplin2008gender, shawcroft2022does, borowsky2000risk}. Non-binary individuals face heightened distress \cite{klinger2024mental} and unique perceptual challenges: because they defy traditional gender expectations, observers may misattribute distress solely to gender identity processes \cite{wall2023trans}. Military veterans are shaped by norms emphasizing toughness and self-reliance, reporting fewer mental health symptoms than civilians despite greater trauma exposure, with stigma as a key barrier \cite{markowitz2023military, hejl2023understanding, sharp2015stigma}. Parents also tend to conceal struggles: mothers suppress distress out of fear of judgment \cite{elliott2020utilization, button2017seeking}, while fathers describe stress-fueled depression centered on inability to cope rather than overt sadness \cite{pedersen2021wanted}.

\paragraph{Rater identity biases and perspectivist annotation.} Psychology research on emotion recognition has long documented an ``in-group advantage'', where individuals identify emotions expressed by members of their own group more accurately than those of outsiders \cite{elfenbein2002universality}. This implies that expressions of distress may be decoded more faithfully by observers who share the expresser's gender, culture, or other identity factors, as they intuitively grasp subtle cues and norms that outsiders overlook. Such premise sets the stage for perspectivist annotation, which incorporates multiple viewpoints (often tied to annotator identities) instead of assuming a single universal ground truth \cite{fleisig2024perspectivist}. Recent work in NLP annotation has begun to embrace this perspectivist philosophy for subjective tasks, such as toxicity and hate-speech detection. \citet{sap2022annotators} showed that annotator identities and beliefs systematically shift toxicity judgments, and \citet{sachdeva2022assessing} found annotators were more likely to label a comment hateful when it targeted their own identity group. Community-perspective datasets have followed, such as \citet{suvarna2025modelcitizens}'s ModelCitizens, demonstrating how in- and out-group annotations surface pluralistic viewpoints that a single aggregated label would erase.

Perspectivist annotation nonetheless remains virtually absent from mental health distress datasets, which typically rely on one undifferentiated set of annotators \cite{shing2018expert, garg2023lost} and collapse their judgments into a single identity-blind label. Given the pronounced socio-demographic differences outlined above, ignoring rater identity risks mislabeling or overlooking distress signals from particular groups, and propagating that bias into downstream mental-health support. Model behavior already hints at this: \citet{gabriel2024can} found GPT-4 inferred demographic characteristics and showed lower empathy toward Black than White posters, and \citet{ma2024evaluating} found LLM chatbots frequently miss LGBTQ-specific nuances and default to risky advice.

\paragraph{Persona-based design for therapeutic alignment.} \citet{song2025typing} define \textit{therapeutic alignment} as grounding AI behavior in users’ \textit{“own definitions of distress and healing”} rather than externally imposed criteria. One strategy is persona-based design, instructing a model to adopt a given identity or role \cite{tseng2024two}. \citet{beck2024sensitivity} find that identity-cued prompts can shift model outputs on subjective tasks, sometimes bringing them closer to subgroup norms. Yet such methods are fragile and can produce stereotyped responses if naively applied: prior work finds demographic prompting only marginally improves accuracy, with gains that are small and context-dependent \cite{hu2024quantifying, beck2024sensitivity}. It is not yet clear whether explicit identity-based prompting can meaningfully mitigate the gap between model defaults and community-specific distress norms. We therefore collect a community-aware dataset that directly tests this question.

\section{Methodology}
\label{method}

\subsection{Dataset}

\paragraph{Data collection and codebook.} For our research, we utilize data from Reddit, which is divided into separate forums known as subreddits. We use publicly available Reddit submissions from an academic torrents dataset derived from Pushshift dumps \cite{baumgartner2020pushshift}. We manually select subreddits that serve as community-oriented discourse spaces associated with specific socio-demographic identities known to vary in their production and perception of psychological distress \cite{de2017gender, klinger2024mental, laws2019differences, treichler2023military}, but not primarily focused on mental health, namely: r/AskMen, r/TwoXChromosomes, r/NonBinary, r/Veterans, r/Mommit, and \mbox{r/daddit}. We treat these subreddits as approximate community contexts rather than exact demographic containers. We consider posts from a 11-year window, specifically from the beginning of 2014 till the end of 2024, and apply a number of structure- and content-based filters. We remove any posts that (i) contain non-textual content (such as images or links), (ii) are stickied or pinned, (iii) contain URLs or mentions of Reddit, (iv) have their content removed or deleted, (v) are less than 10 words (including title), and (vi) received less than 3 comments and a non-positive score\footnote{Doing so removes posts with minimal community engagement, ensuring content reflects active, representative discourse rather than isolated posts.}. While these subreddits cover a wide range of topics relevant to their respective communities, our interest lies in the subset of posts that express psychological distress and seek support. To identify such content, we operationalize distress and support-seeking through the iterative development of an annotation codebook (documented in Appendix~\ref{app:codebook}).

\paragraph{Candidate post identification and purposeful sampling.} 
Since these subreddits are not explicitly mental-health focused, we employed weak supervision to improve the retrieval rate of posts matching our codebook definitions. We used the Snorkel library \cite{ratner2020snorkel} to implement heuristic labeling functions that estimate whether a post contains language associated with psychological distress or support-seeking. We fed keyword lists covering general mental-health terminology, support-seeking phrases, and related emotional-struggle expressions, and retrieved posts in each subreddit that received non-zero probability scores, producing potentially relevant candidates. We then applied an LLM (Llama-3.3:70B) as a pre-screening filter and drew a stratified sample to obtain a balanced set of 200 posts per subreddit (1,200 total) allocating 40 posts to each of five strata: a random stratum drawn from the subreddit before applying weak supervision, plus the four combinations of \(\{\text{mild}, \text{moderate+}\} \times \{\text{seeking}, \text{not\ seeking}\}\). This strategy improves yield for studying distress judgments but creates a {\bf targeted sample} that may over-represent posts with salient distress or support-seeking cues. That could reduce IG--OG differences by making many posts understandable without community context, and it could also amplify LLM over-estimation if models anchor on emotionally loaded language. To assess the sensitivity of our findings to this sampling procedure, we re-estimate the main analyses on the {\bf random stratum}: 40 posts per subreddit drawn before weak supervision and LLM pre-screening. This stratum is more naturalistic with respect to distress-cue selection.

\subsection{Study Design}

\paragraph{Annotation Task.} We recruited U.S.-based adult English speakers via Prolific using identity pre-screening aligned to each subreddit. We define \textbf{in-group (IG)} raters as participants whose self-identified demographics match the focal community of the post's source subreddit, and \textbf{out-group (OG)} raters as those who do not.

Our study uses two distinct annotation conditions, varying the availability of community context across rater types. OG raters were blinded to the post's source community and to their own OG status, providing a post-text-only baseline without community cues. In contrast, IG raters were told which subreddit they were annotating posts from, representing judgments made with relevant community grounding. Accordingly, we interpret any observed IG effect as a contextualized in-group reading effect rather than a pure identity effect, because the IG condition combines demographic matching with subreddit-source disclosure. IG raters also reported subreddit familiarity: across all subreddits, 35\% were at least occasional users and 6\% were active users. Those not already in the community were asked to familiarize themselves by browsing posts and comments for 5--10 minutes.

We use this design because it reflects the comparison central to our study: community-contextualized judgments from in-group readers versus uncontextualized judgments from readers without that grounding. A full $2\times2$ design (IG vs.\ OG $\times$ informed vs.\ blinded) would disentangle identity from source disclosure, a related but distinct question outside our present scope. We also avoid asking OG raters to adopt an insider perspective, since that would measure role-playing rather than naturally grounded interpretation \cite{eyal2018perspective}.

Participants were paid \$10/hour to label 30 posts each. For each post, participants saw the post title and text to answer: (i) distress severity (None/Mild/Moderate+), (ii) support-seeking (Yes/No; asked only if Mild or Moderate+), and (iii) confidence (High/Low). We targeted four IG and four OG ratings per post. After quality-control exclusions (Appendix~\ref{app:annotation_quality}), 9,587 judgments were maintained, resulting in aggregates for 1,198 of the 1,200 sampled posts, from 321 participants. At the end of the task, participants filled in a brief survey, where we prompted them to share their insights and perceptions for the study. We employed the answers from the open-ended questions for our qualitative analysis. The user interface for human annotation, subreddit familiarization, post-task survey, and additional details can be found in the Appendix~\ref{app:screening}.

\subsection{Analytic strategy}
\label{analytic}

\paragraph{Mixed-effect models.} To test our pre-registered hypotheses, we model individual rater judgments using cross-classified mixed-effects logistic regression. Our two binary outcomes are distress $D_{ij}$, coded 1 if rater $i$ labeled post $j$ as ``Mild'' or ``Moderate+'' and 0 if ``None'', and support-seeking $S_{ij}$, coded 1 if rater $i$ indicated the author is seeking support. Each post belongs to one of six author communities (the subreddits) and each rater to a rater group defined by gender, parenthood, and veteran status. Our key predictor is $\mathrm{IG}_{ij}=\allowbreak \text{1}\{\text{rater } i \text{ shares the focal identity of post } j\text{'s community}\}$. For each outcome $y_{ij}\in\{D_{ij}, S_{ij}\}$ we estimate
\[ \operatorname{logit}\!\left(\Pr(y_{ij}=1)\right) = \beta_0 + \beta_{\text{group}[j]} + \gamma_{\text{group}[j]}\cdot \mathrm{IG}_{ij} + u_j + v_i,
\]
where $\beta_{\text{group}[j]}$ is a subreddit fixed effect, $\gamma_{\text{group}[j]}$ is a community-specific IG--OG contrast, and $u_j$, $v_i$ are random intercepts for post and rater. This lets the IG effect vary across communities rather than assuming a single global effect. \textbf{H1a} is evaluated by testing whether the $\gamma_{\text{group}}$ terms jointly differ from zero, via likelihood-ratio tests against a reduced model omitting the $\mathrm{IG}\times\text{community}$ interaction.

For \textbf{H1b} we test whether IG raters agree more than OG raters with their own community's aggregate label. Per-post reference labels are estimated with the Dawid--Skene (DS) model\footnote{DS infers a latent label and rater-specific confusion patterns, down-weighting inconsistent annotators, which suits subjective tasks where rater thresholds vary systematically. Henceforth, ``aggregate'' denotes the DS label derived from at least four IG (or four OG) ratings per post. When the focal rater is IG, we recompute it leave-one-out to avoid circularity.}, yielding the IG and OG aggregates. We then fit the same random-effects structure to the agreement indicator $a_{ij}=\text{1}\{y_{ij}=y_j^{*}\}$, where $y_j^{*}$ is the IG aggregate: first a \emph{global} model with a single pooled IG coefficient $\gamma$ (the overall OR reported below), then an $\mathrm{IG}\times\text{community}$ interaction recovering community-specific contrasts. Full specifications are in Appendix~\ref{app:models}.

\paragraph{Multiple comparisons.} Each interaction model yields six community-specific contrasts per outcome, estimated for two outcomes across two samples, raising a multiplicity concern. We therefore (i) parameterize the models as $y \sim \text{community} + \text{community}\!:\!\mathrm{IG}$ so that each coefficient is the {\bf simple} IG--OG contrast within that community, (ii) treat the six contrasts within each outcome-by-sample combination as a family and report Holm-adjusted $p$-values ($p_{\text{holm}}$) alongside unadjusted ones, and (iii) gate per-community contrasts on that family's omnibus likelihood-ratio test. The pooled $\gamma$ is the pre-registered test and is reported unadjusted.

\paragraph{Qualitative analysis.} To interpret the mechanisms underlying distress perceptions, we conducted an exploratory qualitative analysis of posts that elicited disagreement between the IG and OG aggregates\footnote{We shall refer to these posts as \textit{disputed} posts.}. Detailed methodology and findings are presented in Appendix~\ref{app:qualitative}.

\paragraph{LLM Alignment.} We evaluate how closely different LLMs reproduce each community’s IG judgments, and whether contextualizing the models with group identities and examples improves alignment. The LLM analysis uses the same 1,198 posts. We select Olmo-3-7B, Ministral-3-8B, and Qwen3-30B-A3B, which are competitive open baselines in their size class and provide matched base, instruction-tuned, and reasoning configurations, enabling a controlled ablation of post-training effects on IG alignment. For each model, we evaluate the three variants corresponding to different post-training stages. To test whether the observed patterns generalize beyond mid-sized open-weight models, we additionally evaluate four frontier proprietary configurations: GPT-5 at two reasoning-effort levels (minimal and high), Gemini~2.5~Pro, and Claude~Opus~4. Each model is provided instructions, the codebook, and the post title and text\footnote{Open-weight experiments were run on two NVIDIA A100 80GB GPUs; frontier models were accessed via APIs.}. We query each open-weight variant three times: \textbf{(i) Vanilla ($\varnothing$):} judge the post with task instructions; \textbf{(ii) Contextualized ($C$):} additionally specify the author’s community and ask the model to adopt that community’s perspective; \textbf{(iii) Contextualized with Examples ($CE$):} as in \textbf{(ii)}, but include few-shot examples sampled from disputed posts (randomly selected and shuffled, excluding the target post). Frontier models are evaluated under the vanilla condition only.

Prompt templates and generation examples are in Appendix~\ref{app:prompts}. To assess model-community alignment, we primarily report macro-$F_1$ scores computed against each community's aggregate labels (obtained using the DS model mentioned earlier). Macro-$F_1$ weights classes equally, capturing how well model predictions track a community’s IG labeling behavior. We additionally report inter-annotator reliability using Krippendorff's $\alpha$ in Appendix~\ref{app:agreement} Table~\ref{tab:iar-by-group}.

\section{Results}
\label{results}

Table~\ref{tab:dataset-summary} summarizes the aggregated annotations. IG and OG raters showed moderate agreement on severity ($\kappa = .44$) and substantial agreement on support-seeking ($\kappa = .64$). r/NonBinary had the lowest severity agreement ($\kappa = .39$) but highest support-seeking agreement ($\kappa = .77$), while r/Veterans had the highest severity agreement ($\kappa = .49$). We explore these patterns below. Inter-annotator reliability by group and label type is reported in Appendix~\ref{app:agreement} Table~\ref{tab:iar-by-group}.

\begin{table}
\centering
\small
\setlength{\tabcolsep}{3pt}
\renewcommand{\arraystretch}{1.15}

\resizebox{\columnwidth}{!}{%
\begin{tabular}{@{}l c ccc ccc cc cc@{}}
\toprule
& & \multicolumn{3}{c}{\textbf{IG Severity (\%)}} &
\multicolumn{3}{c}{\textbf{OG Severity (\%)}} &
\multicolumn{2}{c}{\textbf{Seeking (\%)}} &
\multicolumn{2}{c}{\textbf{IG--OG $\kappa$}} \\
\cmidrule(lr){3-5} \cmidrule(lr){6-8} \cmidrule(lr){9-10} \cmidrule(lr){11-12}
\textbf{Community} & \textbf{N} &
\Nsev & \Msev & \Mpsev &
\Nsev & \Msev & \Mpsev &
IG & OG &
Sev. & Seek. \\
\midrule
r/AskMen & 200 & 23.0 & 37.0 & 40.0 & 25.5 & 34.0 & 40.5 & 51.5 & 55.5 & .43 & .56 \\
r/TwoXChro. & 199 & 20.6 & 36.7 & 42.7 & 18.6 & 34.2 & 47.2 & 51.3 & 46.2 & .44 & .68 \\
r/NonBinary & 199 & 21.1 & 42.7 & 36.2 & 22.1 & 43.2 & 34.7 & 45.7 & 48.2 & .39 & .77 \\
r/Veterans & 200 & 31.5 & 35.0 & 33.5 & 25.0 & 41.0 & 34.0 & 43.0 & 47.5 & .49 & .57 \\
r/Mommit & 200 & 21.0 & 34.5 & 44.5 & 25.5 & 39.0 & 35.5 & 45.5 & 44.5 & .43 & .68 \\
r/daddit & 200 & 20.5 & 38.0 & 41.5 & 26.5 & 32.5 & 41.0 & 42.5 & 40.5 & .42 & .57 \\
\midrule
\textbf{Overall} & 1,198 & 23.0 & 37.3 & 39.7 & 23.9 & 37.3 & 38.8 & 46.6 & 47.1 & .44 & .64 \\
\bottomrule
\end{tabular}%
}
\caption{Summary of post-level aggregated annotations showing IG and OG aggregate distributions for severity (Sev.) and support-seeking (Seek.) labels. $N$ is the number of posts with usable aggregates after quality-control exclusions (1,198 of 1,200 sampled). The final two columns report Cohen’s $\kappa$ between the aggregate ratings.}
\label{tab:dataset-summary}
\end{table}

\subsection{RQ1a: Distress distribution differences}
\label{sec:rq1a}

We report results from two complementary analyses: the full purposeful sample (1,198 posts) and the random stratum (240 posts drawn before any weak supervision or LLM pre-screening). Full random-stratum results are in Appendix~\ref{app:random-only} with key findings reported inline below.

\paragraph{Results.} Overall, distress detection rates were very similar across groups (IG: 77.2\%; OG: 77.1\%). The subreddit~$\times$~IG interaction for distress was marginal, $\chi^2(5) = 9.93$, $p = .077$, and the interaction for support-seeking was not significant, $\chi^2(5) = 7.52$, $p = .185$. Neither omnibus test clears conventional significance, we do not treat any full-sample community contrast as evidence of a community-specific effect. For completeness, the largest descriptive contrast was in r/daddit for distress, where IG raters labeled posts as distressed more often than OG raters ($\mathrm{OR} = 2.62$, 95\% CI $[1.18, 5.83]$, unadjusted $p = .018$), but this does not survive correction within its family of six contrasts. No support-seeking contrast was significant after adjustment. Using a stricter ``Moderate+'' only distress threshold and restricting to high-confidence annotations yield the same pattern. \textbf{H1a} is therefore not supported in the full purposeful sample.

\paragraph{Random-stratum analysis.} Since our purposeful sampling targets salient distress cues, the full-sample results above may under-estimate IG--OG divergence. We therefore re-estimated the same models on the random stratum, which reflects the distribution of distress as it naturally occurs in these communities. Under this more naturalistic distribution, both omnibus tests become significant: distress, $\chi^2(5) = 18.07$, $p = .003$; support-seeking, $\chi^2(5) = 13.68$, $p = .018$. The per-community contrasts are therefore interpretable here. For distress, the effect is driven by r/daddit, the only RQ1a contrast that survives Holm correction ($\mathrm{OR} = 7.34$, 95\% CI $[2.43, 22.23]$, $p = .0004$, $p_{\text{holm}} = .003$). For support-seeking, the largest contrast is in r/Veterans and runs in the opposite direction ($\mathrm{OR} = 0.06$, 95\% CI $[0.006, 0.49]$, $p = .009$, $p_{\text{holm}} = .055$), which we discuss below.

\paragraph{Insights.} Taken together, the two analyses tell a coherent story about how sampling interacts with effect detection for \textbf{H1a}. The targeted sample, by design, over-represents posts with salient distress cues that are broadly legible regardless of community membership, compressing IG--OG differences and yielding near-identical detection rates (77.2\% vs.\ 77.1\%). At natural base rates, where subtler expressions of distress predominate, community-specific divergence is substantial: both omnibus interactions become significant, and the largest effect sizes increase several-fold (for distress in r/daddit, from $\mathrm{OR} = 2.62$ to $\mathrm{OR} = 7.34$). Thus, the targeted sample provides a conservative lower bound on IG--OG divergence, while the random stratum reveals the more naturalistic picture. Importantly, the divergence is not a uniform ``insiders see more distress'' effect. It is concentrated in a small number of communities and its direction depends on the outcome, being positive for distress in r/daddit but negative for support-seeking in r/Veterans. OG raters judged r/Veterans posts to be support-seeking substantially more often than IG raters did (86.3\% vs.\ 62.7\%; $\mathrm{OR} = 0.06$, $p = .009$, $p_{\text{holm}} = .055$), consistent with veteran norms of self-reliance and stoicism \cite{markowitz2023military, sharp2015stigma}.

\subsection{RQ1b: IG alignment}
\label{sec:rq1b}

\paragraph{Results.} For distress, raters in the contextualized IG condition showed significantly higher agreement with the aggregate than raters in the uncontextualized OG condition ($\beta = 0.17$, 95\% CI $[0.01, 0.33]$, $p = .035$). This corresponds to $\mathrm{OR} = 1.18$ (95\% CI $[1.01, 1.39]$), where IG raters have about 18\% higher odds of matching the aggregate and a modest difference in raw agreement (IG: 82.3\%; OG: 80.5\%). For support-seeking, there was no evidence of an IG advantage ($\beta = 0.003$, 95\% CI $[-0.17, 0.18]$, $p = .977$; IG: 77.5\%; OG: 77.1\%).

\begin{figure}
\centering
\includegraphics[width=1\linewidth]{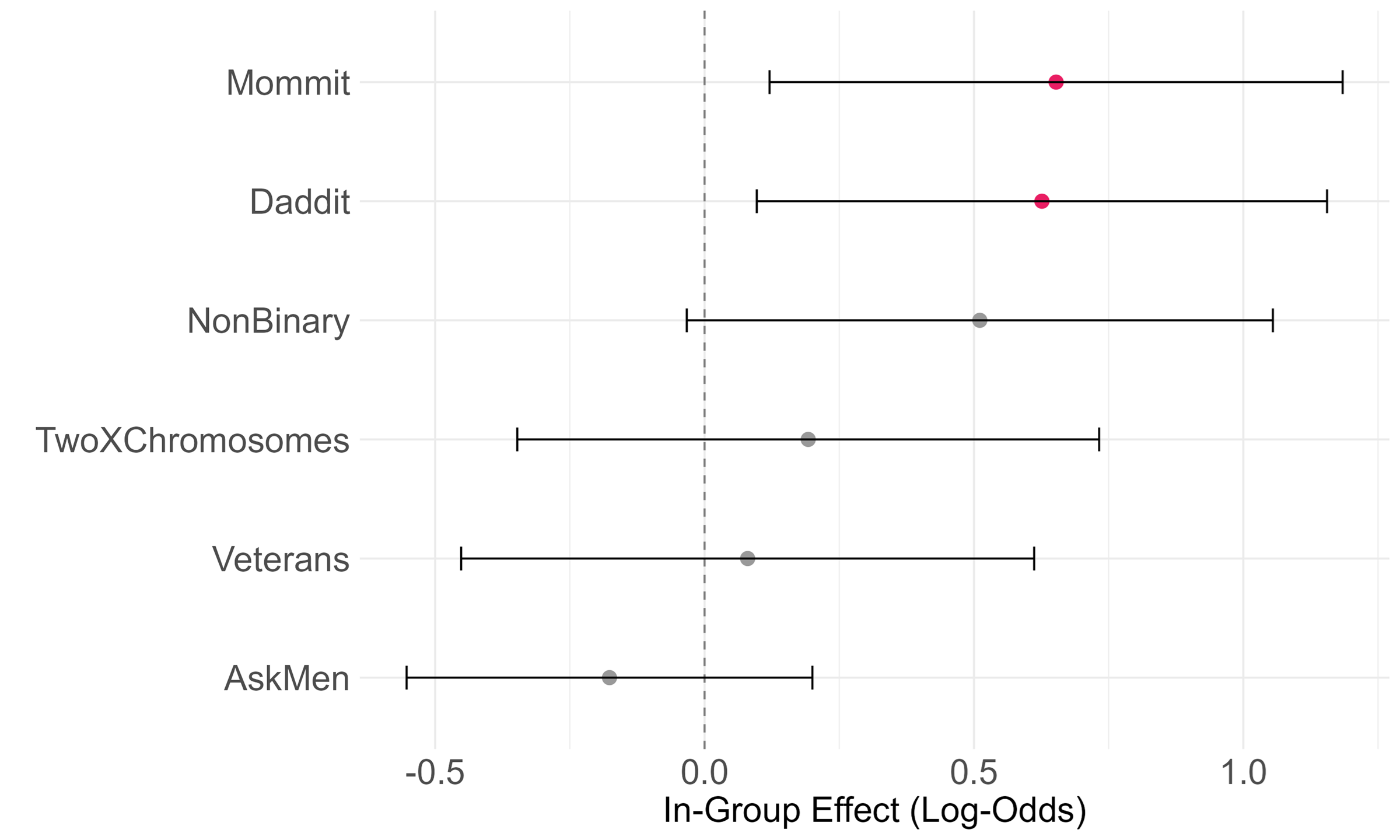}
\caption{IG alignment effect by community; positive values indicate higher IG alignment with the aggregate.}
\label{fig:community_effects}
\end{figure}

Decomposing by community, the omnibus subreddit~$\times$~IG interaction for distress agreement is significant, $\chi^2(5) = 11.10$, $p = .049$, so the per-community contrasts are interpretable, though none survives Holm correction within its family. As shown in Figure~\ref{fig:community_effects}, the effect is concentrated in the parenting communities: r/Mommit ($\mathrm{OR} = 1.61$, 95\% CI $[1.11, 2.34]$, $p = .013$, $p_{\text{holm}} = .078$) and r/daddit ($\mathrm{OR} = 1.57$, 95\% CI $[1.08, 2.27]$, $p = .018$, $p_{\text{holm}} = .088$), with IG raters in these communities having roughly 1.6$\times$ the odds of matching the aggregate. The contrast is weaker in r/NonBinary ($\mathrm{OR} = 1.40$, $p = .095$) and negligible elsewhere, including r/Veterans ($\mathrm{OR} = 0.91$, $p = .615$). For support-seeking agreement the omnibus interaction is far from significant, $\chi^2(5) = 2.67$, $p = .750$, so we report no community contrasts for that outcome.

The random stratum reproduces the pooled effect and slightly strengthens it ($\mathrm{OR} = 1.32$, 95\% CI $[1.01, 1.72]$, $p = .042$), with a strongly significant omnibus interaction, $\chi^2(5) = 21.23$, $p = .0007$. Notably, the single most robust community-level effect anywhere in our analysis emerges from IG raters in r/NonBinary; where they were substantially more likely than OG raters to match their community's distress aggregate ($\mathrm{OR} = 2.96$, 95\% CI $[1.66, 5.29]$, $p = .0002$, $p_{\text{holm}} = .001$). This is consistent with r/NonBinary also showing the lowest IG--OG severity agreement in Table~\ref{tab:dataset-summary} ($\kappa = .39$): it is the community where the two reading conditions diverge most, and correspondingly the one where community grounding matters most.

\paragraph{Insights.} RQ1b offers partial support for our pre-registered hypothesis. Consistent with theories of IG advantages in emotion recognition, raters in the contextualized in-group reading condition were more likely than uncontextualized out-group raters to reproduce their community’s aggregate judgment on distress. Somewhat unexpectedly, however, we did not observe a comparable effect in the r/Veterans community, despite prior work suggesting that shared military experience may facilitate alignment in detecting psychological distress \cite{markowitz2023military}. Additionally, the observed advantage was small overall and did not extend to support-seeking. Thus, \textbf{H1b} is supported for distress but not for support-seeking: the contextualized in-group reading condition is associated with a subtle coordination benefit in interpreting distress signals, rather than the broad IG advantage originally anticipated.

Which communities carry that benefit is not stable across samples. The purposeful sample points to the parenting communities, where neither contrast survives correction, while the random stratum points to r/NonBinary, which does. The defensible claim is therefore the pooled effect together with the significant omnibus heterogeneity, namely that community grounding matters and matters unevenly, rather than that any particular community is the locus of the effect. This motivates RQ2, where we examine whether LLMs differentially align with these aggregates.

\paragraph{Familiarity decomposition.} The IG condition combines shared demographic identity with subreddit-source disclosure, so the design cannot isolate the two. It can, however, test whether the advantage requires deep prior familiarity. We split IG raters into \textbf{IG\textsubscript{NEW}} ($n = 106$), who were not prior subreddit members and received brief familiarization, and \textbf{IG\textsubscript{FAMILIAR}} ($n = 57$), already occasional or active users. Agreement rises from OG (80.5\%) to IG\textsubscript{NEW} (82.1\%) to IG\textsubscript{FAMILIAR} (82.6\%), but almost all of the gain occurs at the OG~$\to$~IG\textsubscript{NEW} step: both subgroups show marginal advantages over OG ($\mathrm{OR} = 1.16$, $p = .097$; $\mathrm{OR} = 1.22$, $p = .075$), yet they do not differ from each other ($\mathrm{OR} = 1.05$, $p = .674$), and the three-level familiarity factor does not improve fit over the binary IG indicator (LRT $\chi^2(1) = 0.18$, $p = .674$). Even brief familiarization thus appears sufficient, and the IG advantage is unlikely to be driven by depth of prior subreddit exposure, though this analysis still cannot separate identity match from source disclosure. Full estimates are in Appendix~\ref{app:familiarity}.

\subsection{RQ2a: Overall LLM--IG alignment}
\label{sec:rq2a}

\paragraph{Analysis.} We evaluate base, instruct, and thinking variants across three model families (Qwen3-30B-A3B, Ministral-3-8B, and Olmo-3-7B) and three prompts ($\varnothing$, $C$, $CE$), alongside three frontier proprietary models under the vanilla condition. We compute macro $F_1$-scores comparing model predictions against IG aggregate labels with bootstrap 95\% confidence intervals. We establish an empirical reference point by computing the median $F_1$ of individual IG raters against the leave-one-out IG aggregate, representing how closely a typical community insider matches their group's aggregate\footnote{Pre-registration note: Support-seeking alignment results are reported as a supplementary outcome in Appendix~\ref{app:llm-ig-seeking}, given its conditional dependence on distress judgments.}.

\begin{figure*}
\centering
\includegraphics[width=1\linewidth]{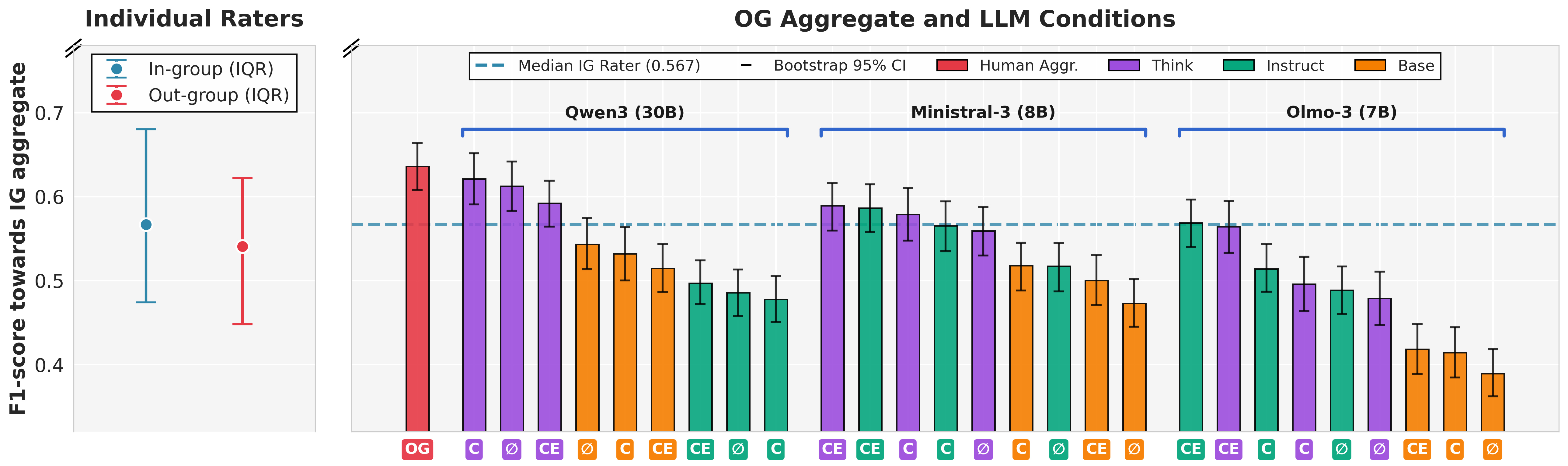}
\caption{Macro $F_1$ score comparison between OG raters and the chosen open-weight LLMs. The dashed line indicates median in-group rater performance, with uncertainty estimated via 95\% bootstrap confidence intervals over the dataset.}
\label{fig:f1-comparison}
\end{figure*}

\paragraph{Results.} Figure~\ref{fig:f1-comparison} presents the main alignment results. Model performance varied substantially by family and post-training stage. Qwen3-30B-A3B Thinking achieved the highest $F_1$ across conditions ($C$: 0.621, +9.6\% vs.\ IG-rater median; $\varnothing$: 0.612, +8.1\%), followed by Ministral-3-8B Thinking ($CE$: 0.589, +4\%) and Ministral-3-8B Instruct ($CE$: 0.586, +3.5\%). In contrast, Olmo-3-7B-Base performed the worst ($\varnothing$: 0.389, $-31.3\%$ vs.\ IG-rater median). Surprisingly, unlike the smaller Ministral-3 and Olmo-3 models, the Instruct variant for Qwen3 under-performed its base counterpart, which may reflect a conservative shift introduced by its instruct tuning. On the other hand, smaller models instruction tuning did improve alignment over their base variants, suggesting miscalibration is not solely a post-training artifact. The topical mixture of pretraining corpora may itself shape how models represent and calibrate the distress severity scale. We probe this hypothesis with a controlled pretraining-mixture ablation in Appendix~\ref{app:datadecide}.

\paragraph{Directional error patterns.} A consistent finding across the open-weight models is the systematic over-estimation of distress. Figure~\ref{fig:directional_error} illustrates how all three open-weight families over-estimate distress severity. Olmo-3 aligned with the IG aggregate on 54\% of posts but over-estimated on 42\% (vs.\ 4\% under-estimation). Ministral-3 showed a similar pattern (59\% aligned, 33\% over, 7\% under), while Qwen3 exhibited the most balanced open-weight profile (62\% aligned, 23\% over, 15\% under). By comparison, OG human aggregates were nearly symmetric (63\% aligned, 18\% over, 19\% under), meaning that this over-estimation cannot be attributed simply to an ``outsider perspective''. Rather, these models behaviorally exhibit an inflated distress prior: they assign higher distress severity than both contextualized in-group aggregates and uncontextualized out-group human judgments. This suggests a calibration bias distinct from lack of community context alone. Our analyses do not isolate a single causal source of this bias; instead, the model-family, post-training-stage, frontier-model, prompting, rationale, and pretraining-mixture analyses point to a multi-stage calibration problem (Appendices~\ref{app:trace-analysis} and~\ref{app:conditional}--\ref{app:datadecide}).

\paragraph{Community-specific bias.} As hypothesized in \textbf{H2a}, error rates varied across communities. Figure~\ref{fig:subreddit_bias} illustrates how posts from r/NonBinary, r/TwoXChromosomes, and r/Veterans exhibited the highest total error, characterized by high over-estimation (40\%--42\%). In contrast, r/AskMen posts showed the smallest error, with the lowest over-estimation (27\%) but the highest under-estimation (12\%). These patterns complement the themes identified in RQ1 (Appendix~\ref{app:qualitative}): communities employing masked communication and baseline calibrations may be misread as elevated distress by observers lacking community context.

\paragraph{Extension to frontier models.} To assess whether the patterns observed in mid-sized open-weight models generalize to state-of-the-art proprietary systems, we additionally evaluate four frontier model configurations under the vanilla prompt condition. Table~\ref{tab:frontier-$F_1$} tabulates macro $F_1$ by community alongside the corresponding open-weight instruct and reasoning variants. Frontier models achieve comparable or slightly higher overall alignment than the best open-weight configuration (Qwen3-30B-A3B Thinking, $F_1 = 0.612$). Full-sample directional error is not uniformly upward, as can be noted from Table~\ref{tab:random-llm-main}. GPT-5 is roughly symmetric (18.6\% over / 19.6\% under at minimal reasoning; 18.4\% / 22.2\% at high), Claude Opus~4 under-estimates (9.6\% / 33.7\%), and only Gemini~2.5~Pro retains a net over-estimation tilt (28.5\% / 9.2\%). What GPT-5 and Gemini do share with the open-weight models is over-estimation at the none-to-mild boundary. Claude Opus~4 served as the exception; even at this boundary it does not show net over-estimation (16\% over / 21\% under). Increased reasoning effort in GPT-5 did not improve alignment ($F_1 = 0.595$ for high vs.\ $0.615$ for minimal) and left none-to-mild over-estimation unchanged, while increasing under-estimation of Moderate+ posts (47\% vs.\ 39\%).

\begin{table}[t]
\centering
\setlength{\tabcolsep}{4pt}
\renewcommand{\arraystretch}{1.15}
\resizebox{\columnwidth}{!}{%
\small
\begin{tabular}{@{}l c c c c c c c@{}}
\toprule
& \textbf{r/AskMen}
& \textbf{r/TwoXChrom.}
& \textbf{r/NonBinary}
& \textbf{r/Veterans}
& \textbf{r/daddit}
& \textbf{r/Mommit}
& \textbf{All} \\
\midrule
\multicolumn{8}{@{}l}{\textit{Open-weight (Instruct)}} \\
\quad Qwen3-30B-A3B      & .600 & .416 & .469 & .461 & .507 & .417 & .485 \\
\quad Ministral-3-8B & .580 & .479 & .505 & .537 & .481 & .492 & .517 \\
\quad Olmo-3-7B      & .582 & .445 & .467 & .416 & .464 & .522 & .488 \\
\midrule
\multicolumn{8}{@{}l}{\textit{Open-weight (Reasoning)}} \\
\quad Qwen3-30B-A3B      & .626 & .524 & .642 & .633 & .592 & .593 & .612 \\
\quad Ministral-3-8B & \textbf{.659} & .481 & .516 & .553 & .552 & .516 & .559 \\
\quad Olmo-3-7B      & .605 & .400 & .431 & .477 & .466 & .424 & .478 \\
\midrule
\multicolumn{8}{@{}l}{\textit{Proprietary}} \\
\quad GPT-5 (minimal)  & .610 & .\textbf{575} & \textbf{.651} & .666 & .578 & .550 & \textbf{.615} \\
\quad GPT-5 (high)     & .603 & .526 & .582 & .637 & .573 & .578 & .595 \\
\quad Gemini 2.5 Pro   & .647 & .549 & .615 & .576 & \textbf{.629} & \textbf{.594} & .609 \\
\quad Claude Opus 4    & .500 & .559 & .569 & \textbf{.672} & .540 & .560 & .573 \\
\midrule
\quad \textit{IG-rater median} & .541 & .529 & .558 & .650 & .588 & .541 & .567 \\
\bottomrule
\end{tabular}%
}
\caption{Macro $F_1$ by community under vanilla prompting. Bold indicates the highest value in each column.}
\label{tab:frontier-$F_1$}
\end{table}

\subsection{RQ2b: Effect of identity-based prompting}
\label{sec:rq2b}

\paragraph{Analysis.} To test whether identity-based prompting improves alignment (\textbf{H2b}), we computed differences in macro $F_1$-scores and examined shifts in error patterns.

\paragraph{Results.} Providing community contextualization and examples improved alignment, particularly for smaller, instruction-tuned models. However, as can be observed in Figure~\ref{fig:f1-comparison}, the magnitude of improvement is modest (typically 2--8 percentage points in $F_1$). For instance, Olmo-3-7B Instruct improved from 0.48 ($\varnothing$) to 0.55 ($CE$), while Ministral-3-8B Instruct improved from 0.52 ($\varnothing$) to 0.59 ($CE$). Larger models such as Qwen3-30B-A3B showed negligible gains, suggesting that contextualization benefits are most pronounced when base capabilities are limited.
Figure~\ref{fig:subreddit_bias} illustrates over- and under-estimation rates across communities under vanilla prompting and with identity-based prompting. Overall, prompting reduced total error for five out of six subreddits, with decreases ranging from 1.0\% (r/Mommit) to 3.5\% (r/Veterans). While contextualization decreased over-estimation rates across all communities, including r/AskMen (from 27\% to 24\%), this came at the expense of increased under-estimation. For r/AskMen, the increase in under-estimation (from 12\% to 18\%) more than offset the decrease in over-estimation, resulting in a net increase in total error. The trade-off was most pronounced in r/TwoXChromosomes, where under-estimation rose by 9\%.

\paragraph{Insights.} These findings provide partial support for \textbf{H2b}: identity-based prompting improves overall alignment with the IG aggregate, but the effect is modest and introduces a systematic shift in error type. Contextualization recalibrates models away from their default over-estimation tendency, but risks pushing them toward under-estimation and missing genuine distress signals.

\paragraph{Exploratory analysis of elicited reasoning traces.} To generate hypotheses about \textit{how} conditioning redistributes errors rather than eliminates them, we conducted a qualitative thematic analysis of elicited model rationales from two reasoning models: Olmo-3-7B-Think and Qwen3-30B-A3B. Given that such outputs are not guaranteed to be faithful accounts of a model's internal decision process \cite{turpin2023language}, the patterns below should be read as exploratory. Full methodology is reported in Appendix~\ref{app:trace-analysis}.

We identified five rationale patterns, two of which dominate vanilla over-estimation: \textit{salient-phrase anchoring} (traces map emotionally loaded phrases to rubric severity thresholds regardless of narrative context) and \textit{impairment over-inference} (traces stretch situational cues into rubric-defined functional impairment without textual support). Conditioning is associated with \textit{stricter rubric enforcement} and \textit{exemplar-driven normalization}, both of which correct over-estimation but also appear in over-correction cases when applied too rigidly. These patterns parallel the human-rater disagreement themes from Appendix~\ref{app:qualitative}, suggesting a structured deficit in the interpretive capacities that distinguish in-group from out-group human raters.

\paragraph{Random-stratum validation.}
\label{sec:random-llm-main}
Given our purposeful sampling targets emotionally loaded posts, the reported over-estimation rates may be inflated. To directly assess this, we re-evaluated all LLM configurations on the random stratum. Table~\ref{tab:random-llm-main} compares over-estimation rates and macro $F_1$ across samples. We find that over-estimation rates are consistently \textit{lower} on the random stratum (e.g., Olmo-3-7B Instruct: 35.9\% vs.\ 41.1\%), as expected if the targeted sample amplifies emotionally salient cues. Overall, over-estimation remains the dominant error direction on the random stratum for the open-weight models and for Gemini~2.5~Pro, GPT-5 is slightly downward (14.3\% over / 20.3\% under) and Claude Opus~4 remains conservative (6.3\% / 30.0\%). Thus, the random-stratum analysis separates magnitude from direction: the targeted pipeline inflates the absolute over-estimation rate, but where models have an upward none-to-mild error, such directional miscalibration is already present in posts sampled before the LLM pre-screening step. 

\section{Discussion and Conclusion}
\label{discussion}

\paragraph{Community differences in distress judgments.} Contrary to \textbf{H1a}, IG and OG raters labeled distress at nearly identical rates in the full purposeful sample, and no full-sample community contrast survives correction. The random stratum indicates this reflects our targeted sampling: at natural base rates the subreddit~$\times$~IG interaction is highly significant and the effects are several-fold larger, so community-specific divergence is more pronounced under naturalistic conditions.

We find modest support for \textbf{H1b}: raters in the contextualized in-group condition were somewhat more likely to match their community's distress aggregate (OR~=~1.18), with significant heterogeneity across communities. The familiarity decomposition shows this advantage is not driven by deep prior subreddit familiarity: IG\textsubscript{NEW} raters with only brief exposure perform comparably to already-familiar IG\textsubscript{FAMILIAR} raters. Since both IG subgroups shared subreddit disclosure and demographic identity, the decomposition rules out a depth-of-familiarity account but cannot separate subreddit disclosure from in-group identity.

Two aspects warrant caution. First, as noted above, which community carries the alignment effect is not stable across samples, so we rest our claims on the pooled effect and the omnibus heterogeneity tests. Second, r/Veterans behaves differently from the rest. The reading-condition effect did not extend to it, plausibly because the ``veteran'' identity spans substantial within-group variation (rank, deployment, combat exposure) that our binary IG/OG distinction cannot capture \cite{britt2020perceived, kulesza2015help}. More strikingly, it is the one community diverging opposite to a simple insider-sensitivity account: out-group raters labeled posts as support-seeking far more often than veterans themselves did, a divergence in labeling rather than in agreement with the aggregate. This fits military norms of self-reliance and stoicism \cite{markowitz2023military, sharp2015stigma}, in which outsiders over-read requests for help into descriptions of difficulty, and bears directly on our central LLM finding by identifying a community where an outsider reading position inflates inferred help-seeking.

\paragraph{Central finding: LLMs have an inflated distress prior.} Open-weight LLMs systematically over-estimate distress relative to both IG and OG judgments. When communities perceive none-to-mild distress, these models are accurate less than half the time, predominantly producing false positives \cite{settanni2025assessing, chim2024overview}. Critically, OG human aggregates show a balanced error profile, whereas the open-weight models over-estimate on 23--42\% of posts but under-estimate on only 4--15\%. Frontier systems do not share this full-sample tilt: Gemini~2.5~Pro still over-estimates, GPT-5 is roughly balanced, and Claude Opus~4 under-estimates. While full-sample over-estimation rates are upper-bound estimates due to targeted sampling, the random-stratum validation confirms that open-weight over-estimation remains the dominant error direction under naturalistic distributions, with 54\% accuracy on none-to-mild posts (vs.\ 44\% on the full sample). As hypothesized in \textbf{H2a}, alignment was weakest for minority communities, with prediction error varying by 6--8 percentage points across groups \cite{santurkar2023whose}.

In line with \textbf{H2b}, identity-based prompting narrowed the performance gap but introduced a systematic shift from over- to under-estimation. Our exploratory reasoning-trace analysis identifies two dominant over-estimation patterns: \textit{salient-phrase anchoring} and \textit{impairment over-inference}, and shows that conditioning corrects these via \textit{stricter rubric enforcement} and \textit{exemplar-driven normalization}. In communities where under-estimation rose substantially (e.g., r/TwoXChromosomes, from 8\% to 17\%), this trade-off could be consequential in real deployments. Persona design is therefore not sufficient on its own.

\paragraph{Implications.} The inflated distress prior has direct consequences for deployment. First, since open-weight models, GPT-5, and Gemini~2.5~Pro over-estimate distress within the none-to-mild boundary, mental health deployments should not treat model severity judgments as well-calibrated point estimates, and instead require calibration and human oversight. On the other hand, conservative systems such as Claude Opus~4 can miss severe cases instead \cite{settanni2025assessing, guo2024large}. Second, community-specific validation is warranted before deployment, particularly for minority populations facing barriers to mental health support \cite{gabriel2024can,ma2024evaluating}. Third, our evidence that conditioning trades over-estimation for under-estimation cautions against naive persona-based design as a bias remedy: sociodemographic prompting is an unstable proxy for subgroup judgment and can worsen alignment \cite{sun2025sociodemographic}. Our elicited-rationale analysis points instead toward calibration on community examples and structured rubric decomposition as targeted post-hoc fixes, while training-time alignment with community judgments \cite{sorensen2024roadmap} addresses the upstream source. ComPO \cite{kumar2025compo} shows that conditioning preference optimization on community identifiers substantially outperforms generic alignment, suggesting that fine-tuning on community data could align severity priors to in-group norms without expert labels. Finally, distress detection risks misuse for surveillance or discriminatory targeting \cite{cosgrove2020digital}, and inferring mental health from social media raises privacy concerns.

\paragraph{Future work.} Three directions follow. First, fine-tuning on community-native interaction data (subreddit comments, replies, voting patterns) should be tested directly against post-hoc contextualization. Second, models should be trained on community-provided rationales rather than labels alone, since the failure modes above are reasoning failures rather than labeling noise. Third, training should model the \textit{distribution} of community judgments rather than collapsing disagreement into a single label.

\paragraph{Limitations.} First, we do not operationalize clinical definitions of distress. We instead capture community perceptions that may diverge from diagnostic criteria, limiting direct comparability with clinical screening tools.

Second, we do not explore diversity within demographic groups. As noted above, identities such as ``veteran'' or ``parent'' encompass substantial internal heterogeneity that our binary IG/OG distinction cannot capture. Subreddit source is likewise an approximate proxy for community context: identity-oriented subreddits may host authors with varying relationships to the focal identity, and their norms may not generalize to the broader demographic group.

Third, our purposeful sampling does not reflect natural base rates, likely compressing IG--OG differences and inflating absolute LLM over-estimation. The random-stratum analyses separate these issues: IG--OG divergence strengthens without salient-cue selection, while for open-weight LLMs absolute over-estimation falls but upward miscalibration persists. The targeted sample should therefore be read as an enriched evaluation set rather than a prevalence estimate.

Fourth, several features of our annotation setup constrain generalizability. Our Prolific pool over-represents English-speaking, educated, technology-engaged US adults \cite{sap2022annotators} who are paid crowd workers rather than organic community members, so the IG advantage likely understates what deeply embedded insiders would show and the true gap between LLMs and community readers may exceed our estimates. Distress and help-seeking norms also vary substantially across cultures \cite{matsumoto1990cultural, de2017gender, rai2025cross}, so alignment patterns may not transfer to non-Western or non-English contexts; Reddit's affordances differ from other platforms where such systems might be deployed; and with only four raters per post the Dawid--Skene aggregates carry non-trivial label uncertainty.

Fifth, we employed an asymmetry in condition framing: OG raters were blinded to the post's source community while IG raters were informed. The RQ1b advantage therefore cannot be attributed to identity alone, and a definitive test would require a design crossing identity with context disclosure.

\paragraph{Conclusion.} Current LLMs are not a drop-in substitute for community judgment of psychological distress. Such judgments are socially situated, and in-group alignment with community aggregates is modest and uneven. Uncontextualized out-group judgments were nearly symmetric, whereas many models misread the lower end of the severity scale. Since frontier systems do not share one overall error direction, deployment cannot assume either a uniformly inflated prior or a uniformly conservative one. The perspectivist contrast between contextualized in-group and uncontextualized outsider reading is what makes that diagnosis possible, though because the in-group condition combines shared identity with subreddit-source disclosure, our human-rater findings are reading-condition effects rather than pure identity effects. Persona conditioning narrows the gap only partially, and at the cost of a compensatory under-estimation risk, which motivates the training-time and evaluation-side remedies outlined above.

\section{Acknowledgements}

The authors would like to acknowledge support from AFOSR, ONR, Minerva, NSF \#2318461, and Pitt Cyber Institute's PCAG awards. The research was partly supported by Pitt’s CRCD resources. Any opinions, findings, and conclusions or recommendations expressed in this material do not necessarily reflect the views of the funding sources.

\bibliography{aaai2027}

\clearpage
\newpage

\subsection{Ethics Checklist}

\begin{enumerate}
\item General items.

\begin{enumerate}
\item Would answering this research question advance science without violating social contracts, such as violating privacy norms, perpetuating unfair profiling, exacerbating the socio-economic divide, or implying disrespect to societies or cultures?
\answerYes{Yes.}
\item Do your main claims in the abstract and introduction accurately reflect the paper's contributions and scope?
\answerYes{Yes. The abstract and introduction clearly enumerate three contributions that match the paper's scope.}
\item Do you clarify how the proposed methodological approach is appropriate for the claims made?
\answerYes{Yes. Section~\ref{method} details the study design, sampling strategy, and analytic approach (mixed-effects models, qualitative analysis) appropriate for the RQs.}
\item Do you clarify what are possible artifacts in the data used, given population-specific distributions?
\answerYes{Yes. We discuss purposeful sampling effects, US-centric annotations, and within-group heterogeneity (e.g., veterans) as potential artifacts.}
\item Did you describe the limitations of your work?
\answerYes{Yes. See Section~\ref{discussion}.}
\item Did you discuss any potential negative societal impacts of your work?
\answerYes{Yes. We discuss risks of LLM over-estimation of distress and implications for equitable AI deployment in mental health contexts.}
\item Did you discuss any potential misuse of your work?
\answerYes{Yes. See Section~\ref{discussion}.}
\item Did you describe steps taken to prevent or mitigate potential negative outcomes of the research, such as data and model documentation, data anonymization, responsible release, access control, and the reproducibility of findings?
\answerYes{Yes. We paraphrase and alter identifying details in excerpts, document prompts and codebook in appendices, and will release code and data upon publication.}
\item Have you read the ethics review guidelines and ensured that your paper conforms to them?
\answerYes{Yes.}
\end{enumerate}

\item Hypotheses testing.
\begin{enumerate}
\item Did you clearly state the assumptions underlying all theoretical results?
\answerYes{Yes. See Section~\ref{intro}.}
\item Have you provided justifications for all theoretical results?
\answerYes{Yes. See Section~\ref{discussion}.}
\item Did you discuss competing hypotheses or theories that might challenge or complement your theoretical results?
\answerYes{Yes. We discuss alternative explanations for null findings (e.g., within-group heterogeneity for veterans, purposeful sampling effects).}
\item Have you considered alternative mechanisms or explanations that might account for the same outcomes observed in your study?
\answerYes{Yes. We discuss alternative mechanisms including purposeful sampling effects, within-group heterogeneity for veterans, model scale/reasoning as mechanisms for bias patterns, and alternative evaluation objectives beyond in-group alignment.}
\item Did you address potential biases or limitations in your theoretical framework?
\answerYes{Yes. See Sections~\ref{results} and~\ref{discussion}.}
\item Have you related your theoretical results to the existing literature in social science?
\answerYes{Yes. We discuss our theoretical foundations in Section~\ref{related-work}.}
\item Did you discuss the implications of your theoretical results for policy, practice, or further research in the social science domain?
\answerYes{Yes. See Section~\ref{discussion}.}
\end{enumerate}

\item Theoretical proofs.

\begin{enumerate}
\item Did you state the full set of assumptions of all theoretical results?
\answerNA{N/A.}
\item Did you include complete proofs of all theoretical results?
\answerNA{N/A.}
\end{enumerate}

\item Machine learning experiments.
\begin{enumerate}
\item Did you include the code, data, and instructions needed to reproduce the main experimental results (either in the supplemental material or as a URL)?
\answerYes{Yes. Prompts and generation parameters are documented in Appendix~\ref{app:prompts}.}
\item Did you specify all the training details (e.g., data splits, hyperparameters, how they were chosen)?
\answerNA{N/A. No training/fine-tuning was performed; we use off-the-shelf models.}
\item Did you report error bars (e.g., with respect to the random seed after running experiments multiple times)?
\answerYes{Yes. See Section~\ref{results}.}
\item Did you include the total amount of compute and the type of resources used (e.g., type of GPUs, internal cluster, or cloud provider)?
\answerYes{Yes.}
\item Do you justify how the proposed evaluation is sufficient and appropriate to the claims made?
\answerYes{Yes. See Section~\ref{analytic}.}
\item Do you discuss what is ``the cost" of misclassification and fault (in)tolerance?
\answerYes{Yes. We discuss how over-estimation may lead to unnecessary interventions and under-estimation may miss genuine distress, with implications for mental health deployments.}

\end{enumerate}

\item Usage of existing assets (e.g., code, data, models) and
release of new assets.
\begin{enumerate}
\item If your work uses existing assets, did you cite the creators?
\answerYes{Yes.}
\item Did you mention the license of the assets?
\answerNA{N/A.}
\item Did you include any new assets in the supplemental material or as a URL?
\answerYes{Yes.}
\item Did you discuss whether and how consent was obtained from people whose data you're using/curating?
\answerYes{Yes. Reddit posts are publicly available; while human annotators provided informed consent via Prolific.}
\item Did you discuss whether the data you are using/curating contains personally identifiable information or offensive content?
\answerYes{Yes. We paraphrase excerpts and alter identifying details. The content warning notes potentially distressing mental health content.}
\item If you are curating or releasing new datasets, did you discuss how you intend to make your datasets FAIR?
\answerNo{No. We need to conform to platforms' policies sharing our curated data. For example, we can only share the public post IDs without the content itself.}
\item If you are curating or releasing new datasets, did you create a Datasheet for the Dataset?
\answerYes{Yes, the datasheet is included with the dataset.}
\end{enumerate}

\item Crowdsourcing and research with human subjects.
\begin{enumerate}
\item Did you include the full text of instructions given to participants and screenshots?
\answerYes{Yes. See Appendix~\ref{app:screening}.}
\item Did you describe any potential participant risks, with mentions of Institutional Review Board (IRB) approvals?
\answerYes{Yes. We included a content warning in our study. Post-task survey results (Figure~\ref{fig:q6}) show participants' comfort levels.}
\item Did you include the estimated hourly wage paid to participants and the total amount spent on participant compensation?
\answerYes{Yes. Participants were paid \$10/hour. In total, we spent \$\textbf{4,197} on participant compensation and platform fees.}
\item Did you discuss how data is stored, shared, and deidentified?
\answerYes{Yes. Reddit excerpts are paraphrased with identifying details altered.}
\end{enumerate}

\end{enumerate}

\clearpage

\section*{Appendix}

\setcounter{section}{0}
\renewcommand{\thesection}{A\arabic{section}}

\setcounter{figure}{0}
\renewcommand{\thefigure}{A\arabic{figure}}

\setcounter{table}{0}
\renewcommand{\thetable}{A\arabic{table}}

\section{Example Posts}\label{app:examples}

Table~\ref{tab:severity-by-group} provides illustrative examples of posts from four communities, alongside the corresponding aggregated IG, OG, and LLM judgments for distress severity and support-seeking. These cases highlight how the three perspectives can diverge on the same post.

\begin{table*}[t]
\centering
\small
\setlength{\tabcolsep}{7pt}
\renewcommand{\arraystretch}{1.2}

\begin{tabularx}{\textwidth}{@{}l >{\raggedright\arraybackslash}X @{\hspace{10pt}} c c c @{}}
\toprule
\textbf{Group} &
\begin{tabular*}{\hsize}{@{\extracolsep{\fill}} l c r @{}}
\textbf{Example} & {\footnotesize \Nsev{} = None \quad \Msev{} = Mild \quad \Mpsev{} = Moderate+ \quad \ssmark{} = Support-seeking} & {}
\end{tabular*}
& \textbf{IG} & \textbf{OG} & \textbf{LLM} \\
\midrule
Mothers &
My six-month-old is getting a precautionary COVID test Tuesday, though it feels like just a cold. She’s been awake since 4am, won’t eat, and I’m spiraling. Too hot, too cold, onesie or pajamas? I’m overwhelmed and don’t know what to do. &
\NsevSS & \MpsevSS & \MpsevSS \\
Fathers &
I have three-year-old twins who share a room; my autistic daughter barely speaks and wakes quietly. One morning she solved a closed door by waking her brother with a toy hammer, using his screams to summon us... I'm impressed with the problem-solving, but terrified at the implications. &
\Mpsev & \Nsev & \Nsev \\
Non-Binary &
Y’all I actually did it: I came out to my parents. The timing sucked, but I did it, I’m not disowned, so that’s something; my emotions are a mess. &
\Nsev & \Msev & \Mpsev \\
Veterans & I drove for 45 minutes, but for about 20 of them I felt completely detached... It was terrifying. I got there, smoked a couple cigarettes, and now I’m scared and confused... what the hell was that?
&
\MpsevSS & \Msev & \MpsevSS \\
\bottomrule
\end{tabularx}

\caption{Paraphrased and shortened examples from the collected Reddit posts and their corresponding aggregated in-group (IG), out-group (OG), and LLM distress and support-seeking judgments.}
\label{tab:severity-by-group}
\end{table*}

\section{Codebook}\label{app:codebook}

We conducted multiple rounds of iterative qualitative coding, each time computing inter-rater reliability and discussing disagreements. The first author and two undergraduate research assistants (one male and one female, both studying Computer Science) collaboratively developed and refined our operational definitions of distress and support-seeking. The team met twice weekly throughout the data analysis period to discuss ambiguities, adjust category definitions, and refine the codebook. Through five iterative rounds of coding and consensus-building, we achieved substantial inter-rater reliability, reaching a Cohen’s $\kappa$ of 0.76 for distress and 0.7 for support-seeking. The final version of the codebook is provided in Figure~\ref{fig:codebook}.

\begin{figure*}[!htb]
\centering
\begin{minipage}{0.98\linewidth}
\begin{mdframed}[
linewidth=0.8pt,
linecolor=black,
innerleftmargin=8pt,
innerrightmargin=8pt,
innertopmargin=8pt,
innerbottommargin=8pt
]
\small
\setlength{\parindent}{0pt}

\noindent
\begin{minipage}[t]{0.49\linewidth}
\vspace{0pt}

\subsection*{1. Emotional or psychological distress severity}

\textbf{Question 1.} \emph{What is the severity level of the author’s self-disclosed emotional and/or psychological distress?}
Select the highest applicable level using the criteria below.

\begin{description}
\item[None]~
\begin{itemize}
\item No first-person (``I'', ``me'', ``my'', etc.) expression of a negative internal state (anxiety, sadness, depression, panic, etc.), or
\item Distress is about someone else, hypothetical, fleeting, or clearly trivial.
\end{itemize}
\textit{Examples:}
\begin{itemize}
\item ``Has anyone tried CBT for panic attacks?''
\item ``Felt a bit sad after that movie.''
\item ``My friend told me she’s depressed, and I’m trying to help her.''
\end{itemize}

\item[Mild]~
\begin{itemize}
\item Negative internal state is explicitly mentioned, but there is no stated functional impairment, or
\item Coping or optimism is visible.
\end{itemize}
\textit{Example:}
\begin{itemize}
\item ``I’m a bit anxious about finals but I’ll be fine after some sleep.''
\end{itemize}

\item[Moderate+]~
\begin{itemize}
\item Negative internal state and at least one sign of interference with daily life (sleep, work, study, relationships), or
\item Explicit hopelessness or worthlessness, or
\item Any suicidal or self-harm ideation, plan, recent attempt, or intent.
\end{itemize}
\textit{Examples:}
\begin{itemize}
\item ``For weeks I can’t concentrate at work because I’m so anxious.''
\item ``Keep waking up at 3 a.m. with panic.''
\item ``My depression makes me feel worthless and I’m failing all my courses.''
\item ``Nothing matters anymore, I skip classes most days.''
\item ``Thinking of ending it.''
\end{itemize}
\end{description}

\end{minipage}\hfill
\begin{minipage}[t]{0.49\linewidth}
\vspace{0pt}

\subsection*{2. Support seeking}

\textbf{Question 2.} \emph{Is the author of the post seeking support about their distress?} (This question is only asked if the post’s severity level is \textbf{Mild} or \textbf{Moderate+}.) Answer \textbf{YES} if the post contains \emph{all} of the following:

\begin{enumerate}
\item \textbf{Support-seeking cue:}
\begin{itemize}
\item The post either contains:
\begin{itemize}
\item A clear appeal for social support \\
(e.g., ``Can anyone relate \dots?'', ``Can someone suggest how I can \dots?'', ``I need \dots'', ``Please help me \dots'', ``Does anyone have suggestions \dots'', etc.), \textbf{or}
\item An implicit appeal, where the author signals a desire for engagement or support \\
(e.g., ``I don’t know if I’m looking for advice or what \dots'', ``I feel like I have no support'', ``I’m really struggling and don’t know who to talk to'').
\end{itemize}
\end{itemize}

\item \textbf{Distress-focused:}
\begin{itemize}
\item The request for support is centered around the situation or feelings the author describes.
\end{itemize}

\item \textbf{Author-centered:}
\begin{itemize}
\item The author is seeking support for themselves, not just starting a conversation or making a rhetorical point.
\end{itemize}
\end{enumerate}

\textit{Examples of posts \textbf{seeking} support:}
\begin{itemize}
\item ``My new SSRI has left me nauseous and exhausted. Has anyone figured out how to manage these side effects without switching meds?''
\item ``I don’t know if I’m looking for advice or what. I just feel like I have no support anywhere on this.''
\item ``I’m ready to tell my parents about my depression but don’t know where to start. Any advice on opening that conversation?''
\end{itemize}

\textit{Examples of posts \textbf{not seeking} support:}
\begin{itemize}
\item ``I hate everything. Just needed to get that off my chest.'' (Makes no request or invitation for engagement.)
\item ``Does anyone else ever feel anxious before a big meeting?'' (Invites poll/discussion, not seeking feedback related to a personal situation.)
\item ``It’s Friday and I’m drowning my feelings before I head downtown. What about you guys?'' (Small-talk invitation, not an ask for help.)
\end{itemize}

\end{minipage}

\end{mdframed}
\end{minipage}

\caption{Annotation codebook.}
\label{fig:codebook}
\end{figure*}

\section{Annotation task}\label{app:screening}

\paragraph{Participant Screening.} Participants were partitioned into 14 demographically defined rater groups based on whether their self-identified demographic characteristics aligned with those of the Reddit forum targeted by each post. Under this definition, the six IG groups represented men, women, non-binary individuals, veterans, mothers, and fathers, with the remaining eight groups serving as out-group comparison arms. For identities defined by multiple attributes, Prolific’s pre-screening does not support expressing the full OG as a single audience filter. We therefore recruited OG raters via separate arms, which explains why there were eight out-group arms. Each subreddit received approximately 1,600 ratings: 800 in-group and 800 out-group, totaling 9,600 across six subreddits.

\begin{itemize}
\item r/AskMen: 800 men (in-group), 800 not-men (out-group).
\item r/TwoXChromosomes: 800 women, 800 not-women.
\item r/NonBinary: 800 non-binary, 800 not-non-binary.
\item r/Veterans: 800 veterans, 800 non-veterans.
\item r/daddit: 800 male parents; out-group split between not-men ($x$) and not-parents ($800-x$).
\item r/Mommit: 800 female parents; out-group split between not-women ($y$) and not-parents ($800-y$).
\end{itemize}

Here, $x, y \in [0, 800]$; the two parenting communities' out-group allocations are intentionally flexible so we can balance supply between ``non-target-gender'' raters and ``non-parent'' raters while keeping each parenting subreddit at 800 out-group ratings. These allocations correspond to 11 prescreened arms on Prolific: six in-group arms (Male; Female; Non-Binary; Veteran = Yes; Parent = Yes \& Male; Parent = Yes \& Female) and five out-group arms (NOT Male; NOT Female; NOT Non-Binary; Veteran = No; Parent = No). Table~\ref{table:prolific_data} tabulates a summary of the study's participants' demographic information.

\paragraph{Study Flow.} Algorithm~\ref{alg:flow} summarizes the flow for any participant in our study. As outlined, eligible Prolific participants enter via a demographically filtered study link, provide consent, pass comprehension checks, are placed into the corresponding rater group, contextualized within the subreddit if they are in-group raters and they are not already familiar, and are then randomly assigned a 30-post bundle drawn from the pre-shuffled slot queue. While participants were assigned 30 posts each, the actual number of ratings per participant varied slightly due to variations in task completion.

\begin{algorithm}[t]
\DontPrintSemicolon
\caption{Participant flow throughout the study}\label{alg:flow}
\KwIn{Candidate participant $p$}
\KwOut{Responses for a 30-post bundle (or partial), and payment status}

\While{study quotas not met}{
Recruit candidate $p$\;

\If{\textbf{not}(\textsc{US}($p$) \textbf{and} \textsc{Age}($p$)\,$\ge 18$ \textbf{and} \textsc{English}($p$))}{
\textbf{continue}\;
}

$(\textit{arm}, \textit{Subreddit}) \leftarrow \textsc{AssignByDemographics}($p$)$\;

\If{\textbf{not} \textsc{ConsentAndComprehension}($p$)}{
\textsc{ReplaceParticipant}()\;
\textbf{continue}\;
}

\If{\textit{arm} = \textsc{InGroup}}{
\textsc{SubredditFamiliarization}($p$,\textit{Subreddit})\;
}

\textit{bundle} $\leftarrow$ \textsc{Assign30PostsRandomOrder}(\textit{Subreddit})\;
\textit{completed} $\leftarrow$ \textsc{CollectResponses}($p$,\textit{bundle})\;

\If{\textsc{FailAICheck}($p$) \textbf{or} \textsc{ExitedEarly}($p$)}{
\textsc{PayForCompletedPosts}($p$,$\lvert \textit{completed}\rvert$)\;

\textsc{ReplaceParticipant}()\;
\textbf{continue}\;
}

\textsc{PayForBundle}($p$,30)\;
\textbf{break}\;
}
\end{algorithm}

\begin{figure*}
\centering
\includegraphics[width=1\linewidth]{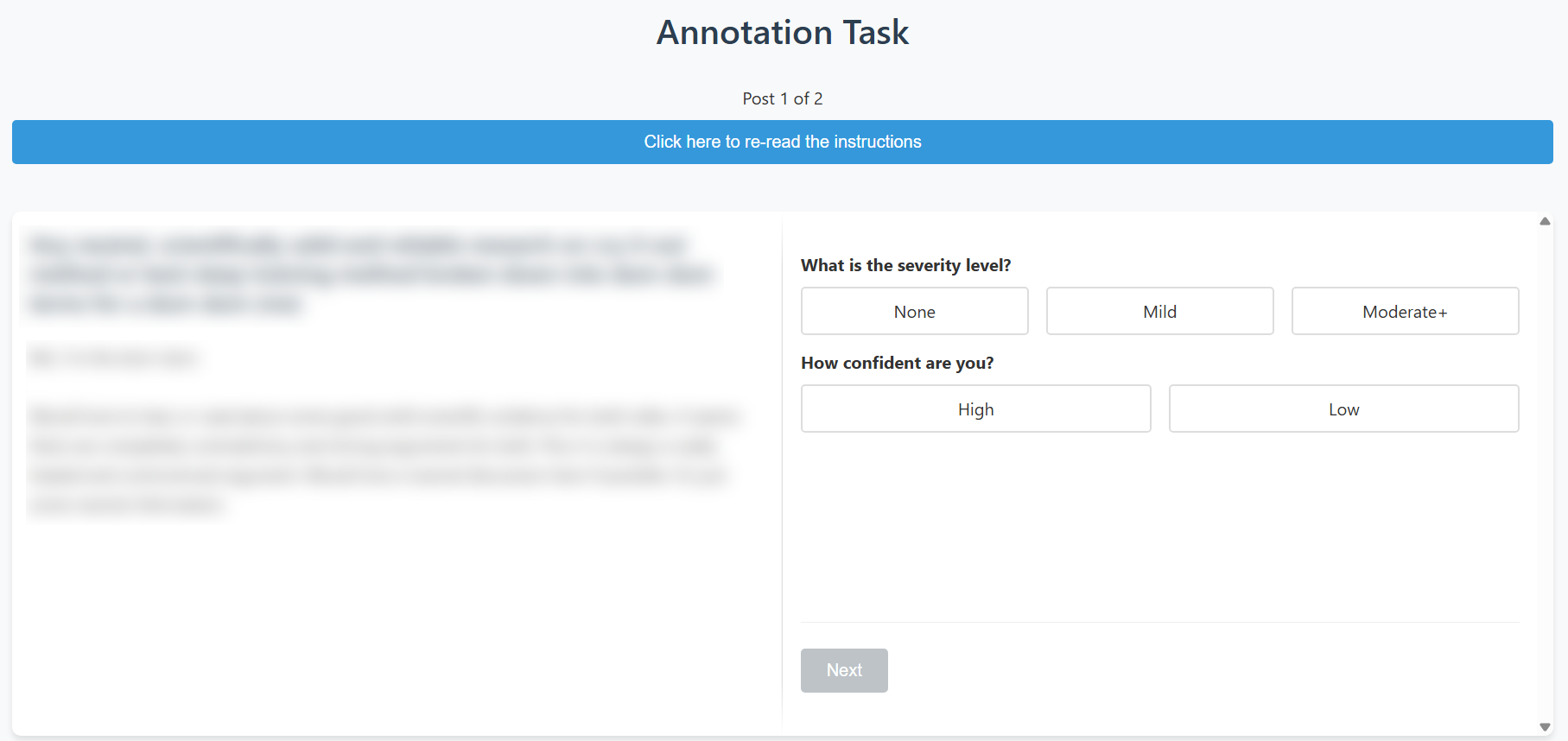}
\caption{Annotation Task interface.}
\label{fig:task}
\end{figure*}

\begin{figure*}
\centering
\includegraphics[width=1\linewidth]{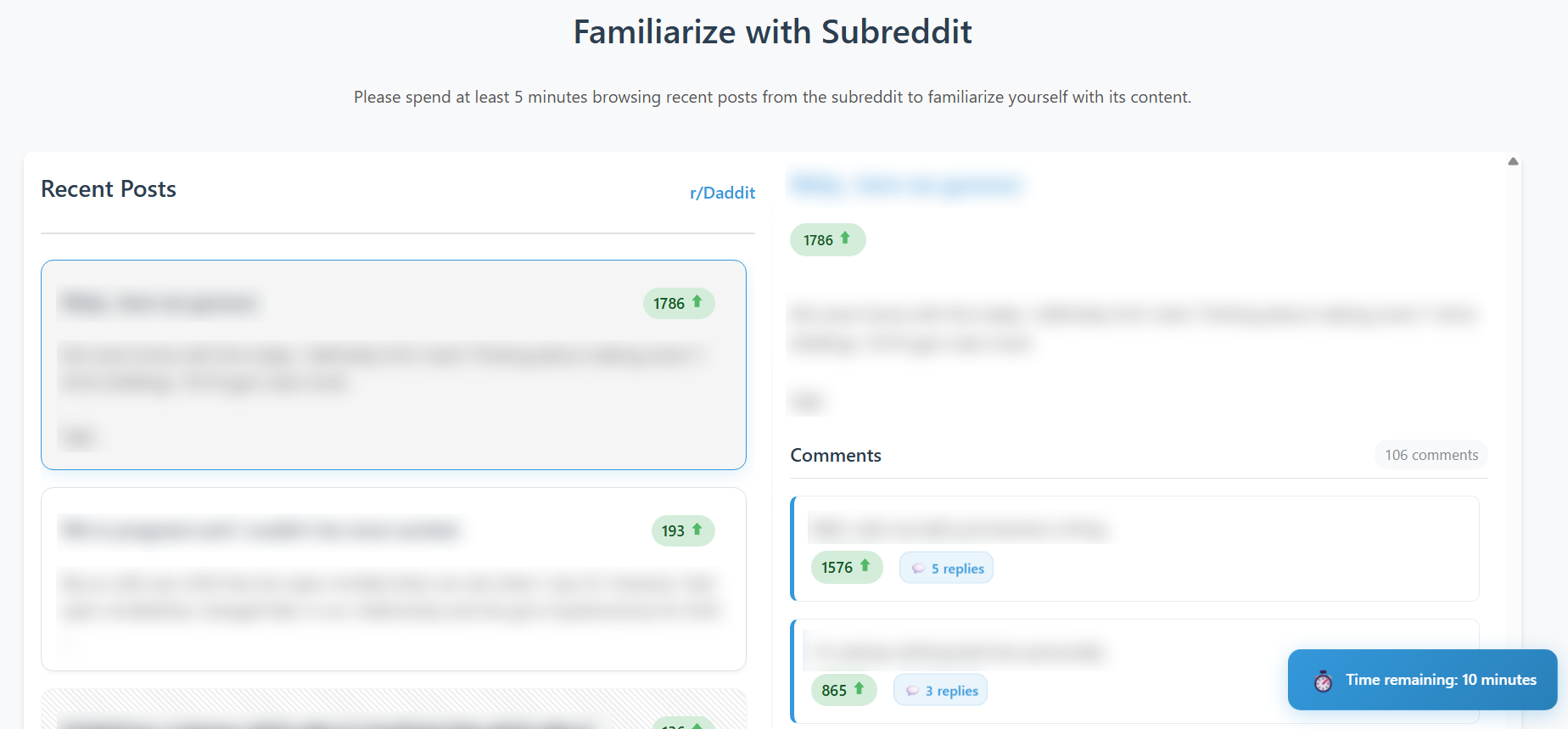}
\caption{Subreddit Familiarization interface (for participants assigned to an in-group arm who were not occasional or active members of their assigned subreddit).}
\label{fig:fam}
\end{figure*}

\paragraph{Annotation Interface.} The interfaces used for the task and the subreddit familiarization are provided in Figure~\ref{fig:task} and Figure~\ref{fig:fam}.

\begin{table*}[p]
\centering
\scriptsize
\setlength{\tabcolsep}{2.5pt}
\renewcommand{\arraystretch}{1.15}
\caption{Demographic summary by subreddit and group membership. Data for \textit{Gender}, \textit{Children}, and \textit{Military veteran} were only provided whenever their pre-screening was required for eligibility. Some demographic responses may also be missing because participants did not provide consent to share the information on Prolific.}
\label{table:prolific_data}
\begin{tabularx}{\textwidth}{
>{\raggedright\arraybackslash}p{2.2cm}
>{\centering\arraybackslash}p{0.6cm}
>{\raggedright\arraybackslash}p{1.3cm}
>{\raggedright\arraybackslash}X
>{\raggedright\arraybackslash}X
>{\raggedright\arraybackslash}p{0.9cm}
>{\raggedright\arraybackslash}X
>{\raggedright\arraybackslash}X
>{\centering\arraybackslash}p{0.8cm}
>{\raggedright\arraybackslash}p{1.4cm}
}

\toprule
\textbf{Subreddit} & \textbf{In-Group} & \textbf{Age} & \textbf{Employment status} & \textbf{Ethnicity} & \textbf{Student status} & \textbf{Sex} & \textbf{Gender} & \textbf{Children} & \textbf{Military veteran} \\
\midrule

r/AskMen & No & Mean: 42.7, Range: 22--73 &
Full-Time (7), Other (2), Part-Time (2), Not in paid work (1) &
White (16), Black (7), Mixed (2), Asian (1), Other (1) &
No (9), Yes (2) &
Female (27) &
Woman (26), Non-binary (1) &
\diagempty &
\diagempty \\

r/AskMen & Yes & Mean: 38.2, Range: 20--62 &
Full-Time (11), Part-Time (2), Unemployed (1), Not in paid work (1) &
White (17), Mixed (4), Black (2), Other (2), Asian (1) &
No (10), Yes (2) &
Male (27) &
Man (27) &
\diagempty &
\diagempty \\

r/daddit & No & Mean: 39.6, Range: 22--78 &
Full-Time (4), Part-Time (2), Other (1) &
White (10), Black (6), Asian (3), Mixed (1) &
No (8), Yes (1) &
Female (18), Male (2) &
Woman (12) &
No (8) &
\diagempty \\

r/daddit & Yes & Mean: 40.6, Range: 24--65 &
Full-Time (7), Part-Time (3) &
White (20), Black (6), Mixed (1) &
No (10), Yes (3) &
Male (27) &
Man (27) &
Yes (27) &
\diagempty \\

r/Mommit & No & Mean: 38.1, Range: 23--53 &
Full-Time (9), Unemployed (2), Not in paid work (2), Part-Time (1) &
White (16), Black (5), Asian (1), Mixed (1) &
No (13), Yes (1) &
Male (16), Female (7) &
Man (12) &
No (11) &
\diagempty \\

r/Mommit & Yes & Mean: 48.8, Range: 29--79 &
Full-Time (10), Other (2), Not in paid work (2), Unemployed (1), Part-Time (1) &
White (18), Black (6), Mixed (2), Asian (1) &
No (11), Yes (2) &
Female (27) &
Woman (27) &
Yes (27) &
\diagempty \\

r/NonBinary & No & Mean: 40.0, Range: 19--70 &
Full-Time (10), Part-Time (3), Not in paid work (1) &
White (18), Asian (3), Mixed (3), Black (2), Other (1) &
No (13), Yes (1) &
Female (14), Male (13) &
Woman (15), Man (12) &
\diagempty &
\diagempty \\

r/NonBinary & Yes & Mean: 31.8, Range: 20--50 &
Part-Time (8), Full-Time (6), Unemployed (3), Other (1), Not in paid work (1) &
White (15), Mixed (6), Black (3), Other (3), Asian (1) &
No (14), Yes (5) &
Female (22), Male (5), Prefer not to say (1) &
Non-binary (28) &
\diagempty &
\diagempty \\

r/TwoXChromosomes & No & Mean: 40.2, Range: 21--72 &
Full-Time (14), Part-Time (1), Unemployed (1), Not in paid work (1) &
White (18), Black (4), Asian (3), Other (1), Mixed (1) &
No (12), Yes (5) &
Male (27) &
Man (26), Non-binary (1) &
\diagempty &
\diagempty \\

r/TwoXChromosomes & Yes & Mean: 41.4, Range: 21--63 &
Full-Time (9), Part-Time (2), Unemployed (2), Other (1), Not in paid work (1) &
White (22), Asian (2), Black (2), Mixed (1) &
No (12), Yes (3) &
Female (27) &
Woman (27) &
\diagempty &
\diagempty \\

r/Veterans & No & Mean: 44.2, Range: 19--80 &
Full-Time (4), Unemployed (3), Part-Time (3), Not in paid work (1), Other (1) &
White (7), Black (6), Mixed (4), Other (1) &
No (9), Yes (3) &
Female (12), Male (6) &
\diagempty &
\diagempty &
No (18) \\

r/Veterans & Yes & Mean: 45.8, Range: 29--66 &
Full-Time (8), Not in paid work (3), Part-Time (2) &
White (19), Mixed (3), Black (3), Asian (1) &
No (13), Yes (1) &
Male (14), Female (13) &
\diagempty &
\diagempty &
I'm a US services veteran (27) \\

\midrule
\textbf{Total} & & Mean: 40.9, Range: 19--80 &
Full-Time (99), Part-Time (30), Not in paid work (14), Unemployed (13), Other (8) &
White (196), Black (52), Mixed (29), Asian (17), Other (9) &
No (134), Yes (29) &
Female (167), Male (137), Prefer not to say (1) &
Woman (107), Man (104), Non-binary (30) &
Yes (54), No (19) &
I'm a US services veteran (27), No (18) \\

\bottomrule
\end{tabularx}
\end{table*}

\paragraph{Quality Assurance and Integrity Safeguards.}
\label{app:annotation_quality}

We implemented quality-control measures to ensure participants understood the codebook and produced attentive, genuinely human annotations.

\begin{itemize}
\item \textbf{Comprehension checks:} Before starting the main task, participants reviewed the codebook and completed a practice round consisting of ten posts. These items were sampled from our earlier qualitative phase (detailed in Section~\ref{app:codebook}), selected specifically because of high confidence across the three annotators. Participants who disagreed with the reference labels on more than three out of ten items did not proceed to the main study.

\item \textbf{Attention checks:} To identify inattentive annotations, we embedded three explicit attention-check items within each bundle of 30 posts. These items clearly instructed the participant which label to select for the presented post. If a participant failed two out of the three attention checks within a bundle, we excluded their annotations from analysis.

\item \textbf{Safeguards against AI-assisted responses:} Recent evidence suggests that crowd workers may use LLMs to increase speed and throughput in annotation tasks, compromising the validity of crowd-sourced data \cite{veselovsky2023artificial}. In response, we implemented multiple preventive and detection-oriented safeguards:

\begin{itemize}
\item \textbf{Explicit prohibition.} Our study instructions explicitly asked participants not to use AI while completing the task.
\item \textbf{Blocking browser-based agents.} We restricted access from known browser-based agent environments to reduce automated or semi-automated completion.
\item \textbf{Flagging selection and copying.} Inspired by the aforementioned work using behavioral signals to study LLM usage in crowdsourcing, we incorporated a check within the annotation interface to flag participants who selected and copied text for a given item. Participants flagged during the comprehension check did not proceed to the main task. Participants flagged during the task itself had their corresponding flagged annotations excluded from the final dataset. In total, \textbf{38\% of participants} who began the study were flagged at the screening stage.
\end{itemize}
\end{itemize}

While no protocol can fully eliminate the possibility of participants employing AI throughout their participation, we view these safeguards as essential for improving annotation reliability and ensuring that our dataset reflects genuine human judgments.

After applying these exclusions, we retained a post only if it still had at least four IG and four OG ratings. Two posts (specifically, one from r/TwoXChromosomes and one from r/NonBinary) fell below this threshold and were dropped, leaving 9,587 judgments.

\section{Post-task survey}

After completing the annotation task, participants completed a short survey, where we elicited their perceptions of the study itself.

\paragraph{Agreement statements.} Participants indicated their agreement on a 5-point Likert scale (Strongly Disagree - Strongly Agree) with the following statements:
\begin{enumerate}
\item The task was mentally demanding.
\item I had to invest a lot of effort to do the annotations well.
\item The labeling instructions were clear.
\item I am confident I understand the subreddit's core norms (topic focus, tone, what is on- vs.\ off-topic) \textbf{(shown to IG participants only)}.
\end{enumerate}

\paragraph{Task reflection questions.}
\begin{enumerate}
\setcounter{enumi}{4}
\item Which part of the label was hardest?
\begin{itemize}
\item Distress severity (None, Mild, Moderate+)
\item Support-seeking
\item Both equally
\item Neither, easy overall
\end{itemize}

\item Did any post make you feel distressed or uncomfortable?
\begin{itemize}
\item Extremely uncomfortable
\item Somewhat uncomfortable
\item Not at all
\end{itemize}

\item Were there any parts of the task that were unclear, unhelpful, or frustrating? If so, which part and why? Do you have suggestions to improve the instructions or the interface? \textbf{(open-ended question)}

\item Thinking about the examples you rated low confidence: in a few sentences, what factors made you feel uncertain about your answers? \textbf{(open-ended question)}
\end{enumerate}

\begin{figure*}[p] 
\centering

\begin{subfigure}[t]{0.49\linewidth} 
\centering
\includegraphics[width=\linewidth]{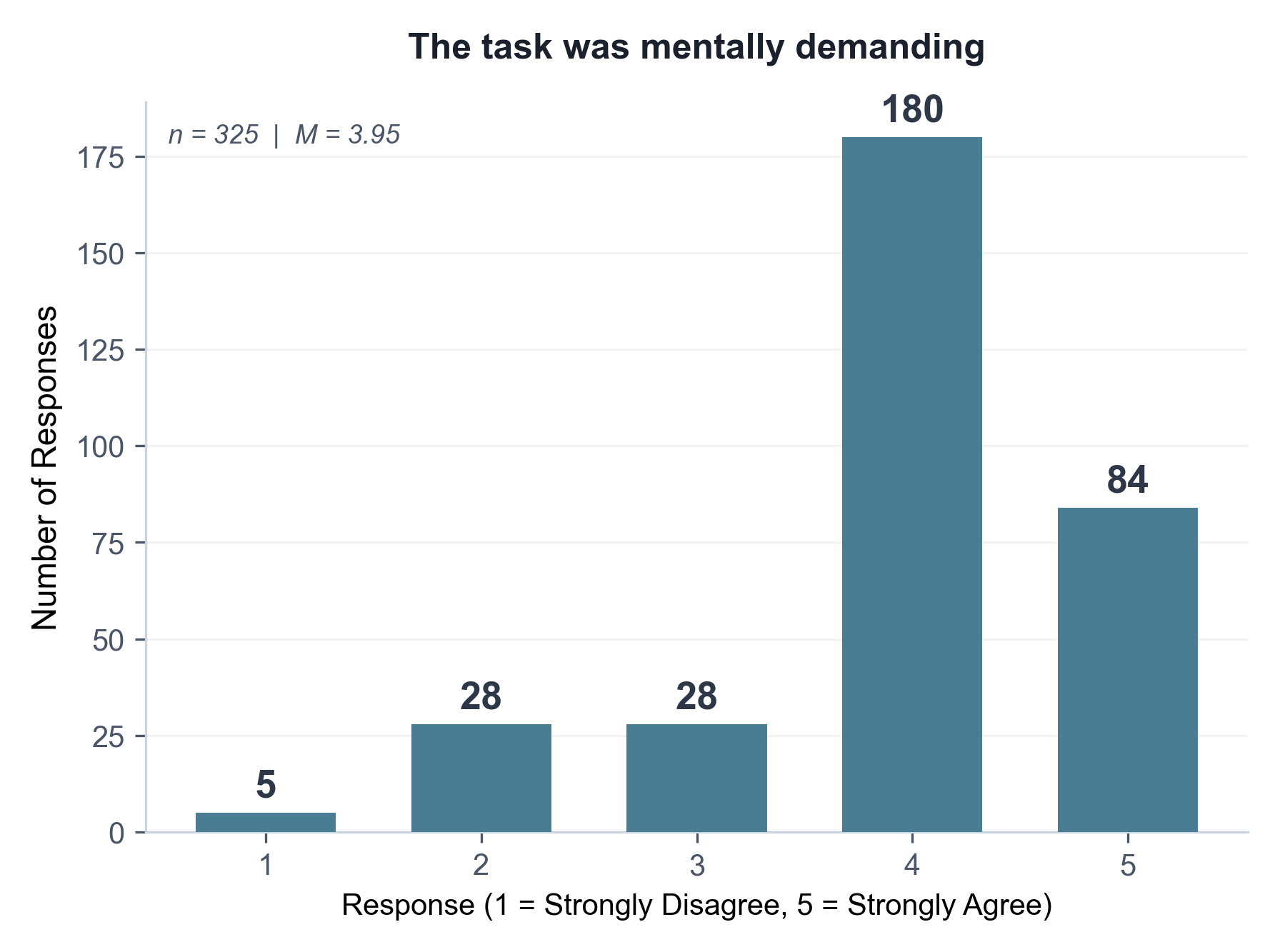}
\caption{Question 1.}
\label{fig:q1}
\end{subfigure}\hfill
\begin{subfigure}[t]{0.49\linewidth}
\centering
\includegraphics[width=\linewidth]{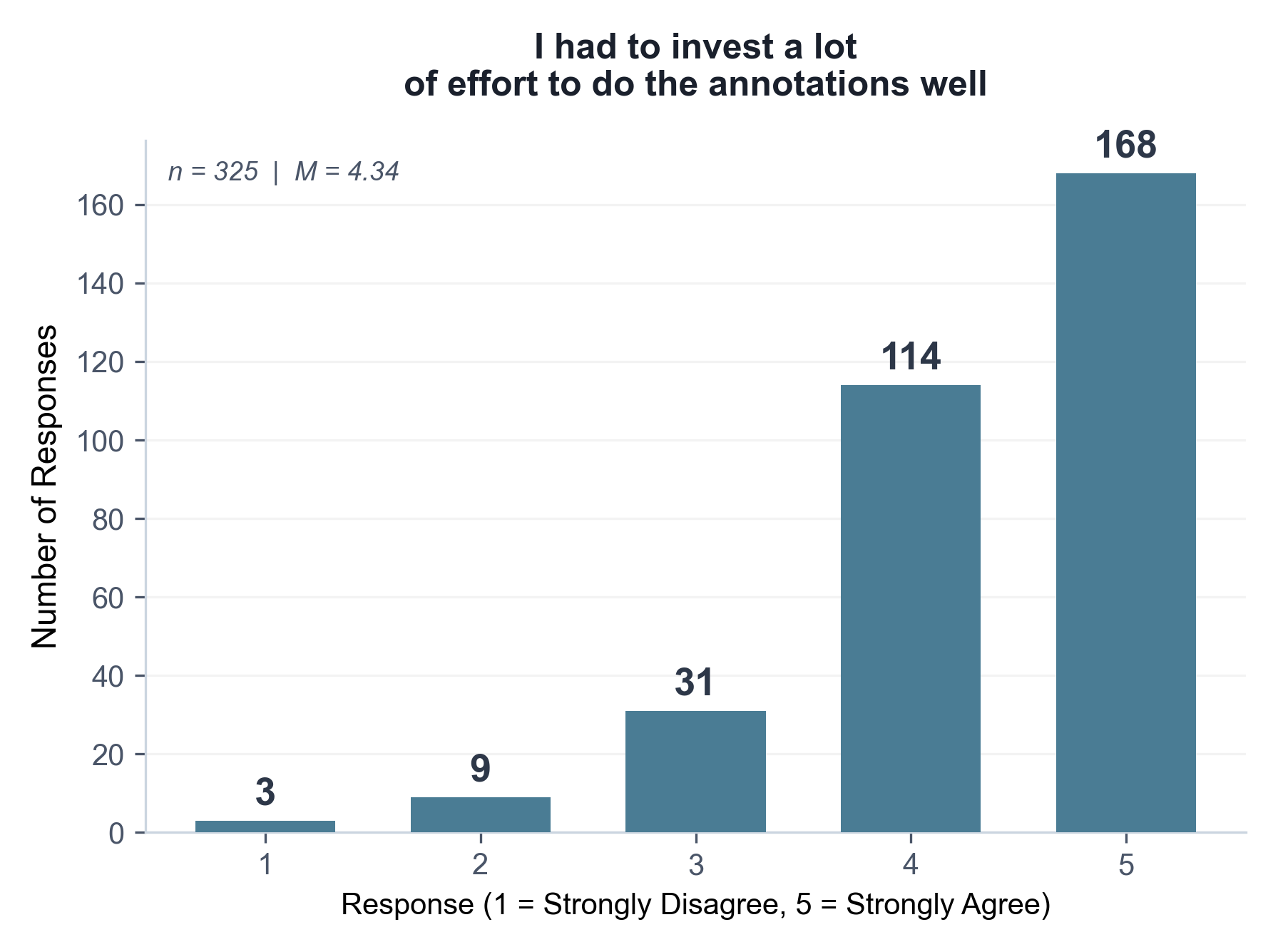}
\caption{Question 2.}
\label{fig:q2}
\end{subfigure}

\vspace{0.6em}

\begin{subfigure}[t]{0.49\linewidth}
\centering
\includegraphics[width=\linewidth]{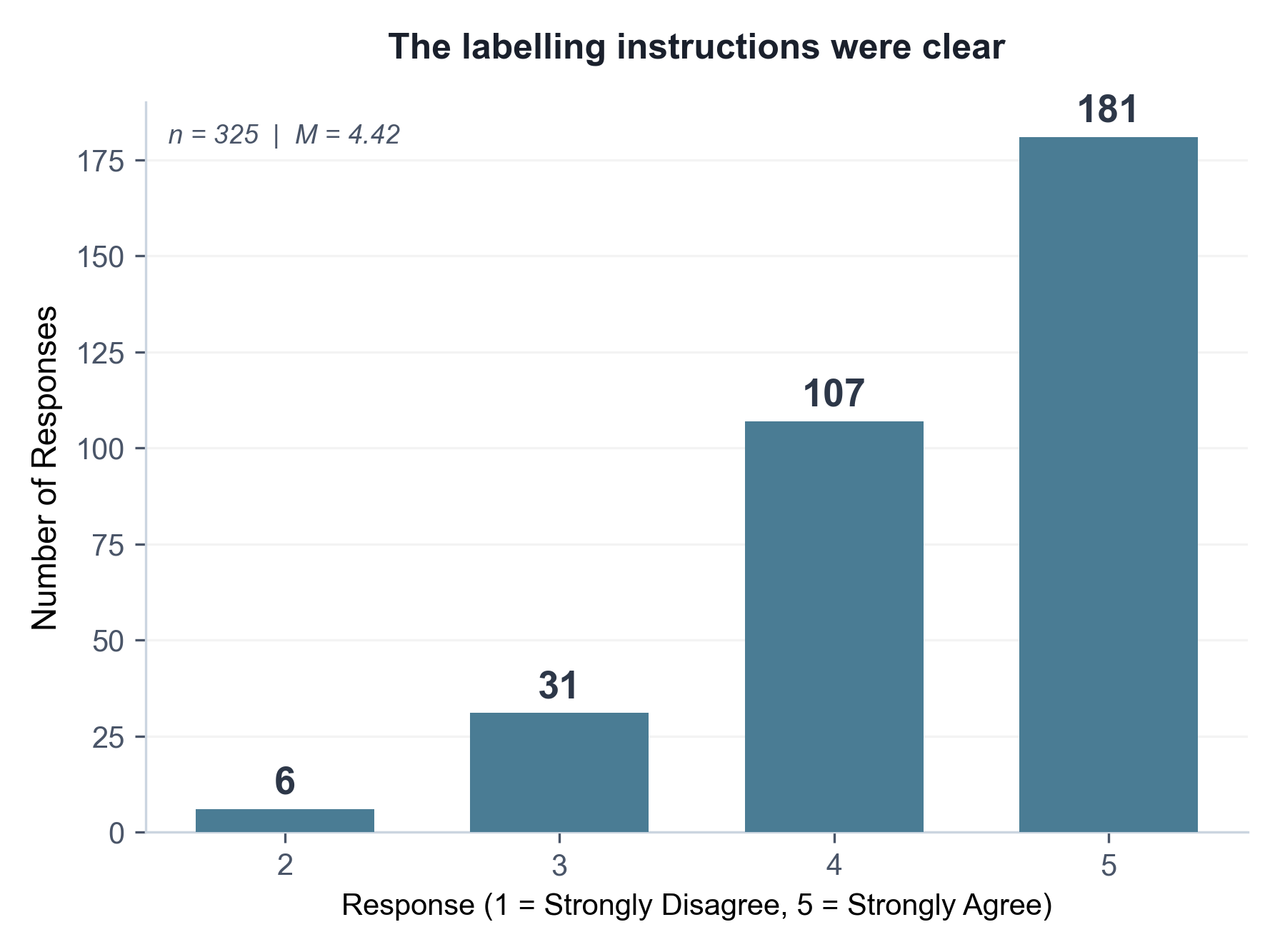}
\caption{Question 3.}
\label{fig:q3}
\end{subfigure}\hfill
\begin{subfigure}[t]{0.49\linewidth}
\centering
\includegraphics[width=\linewidth]{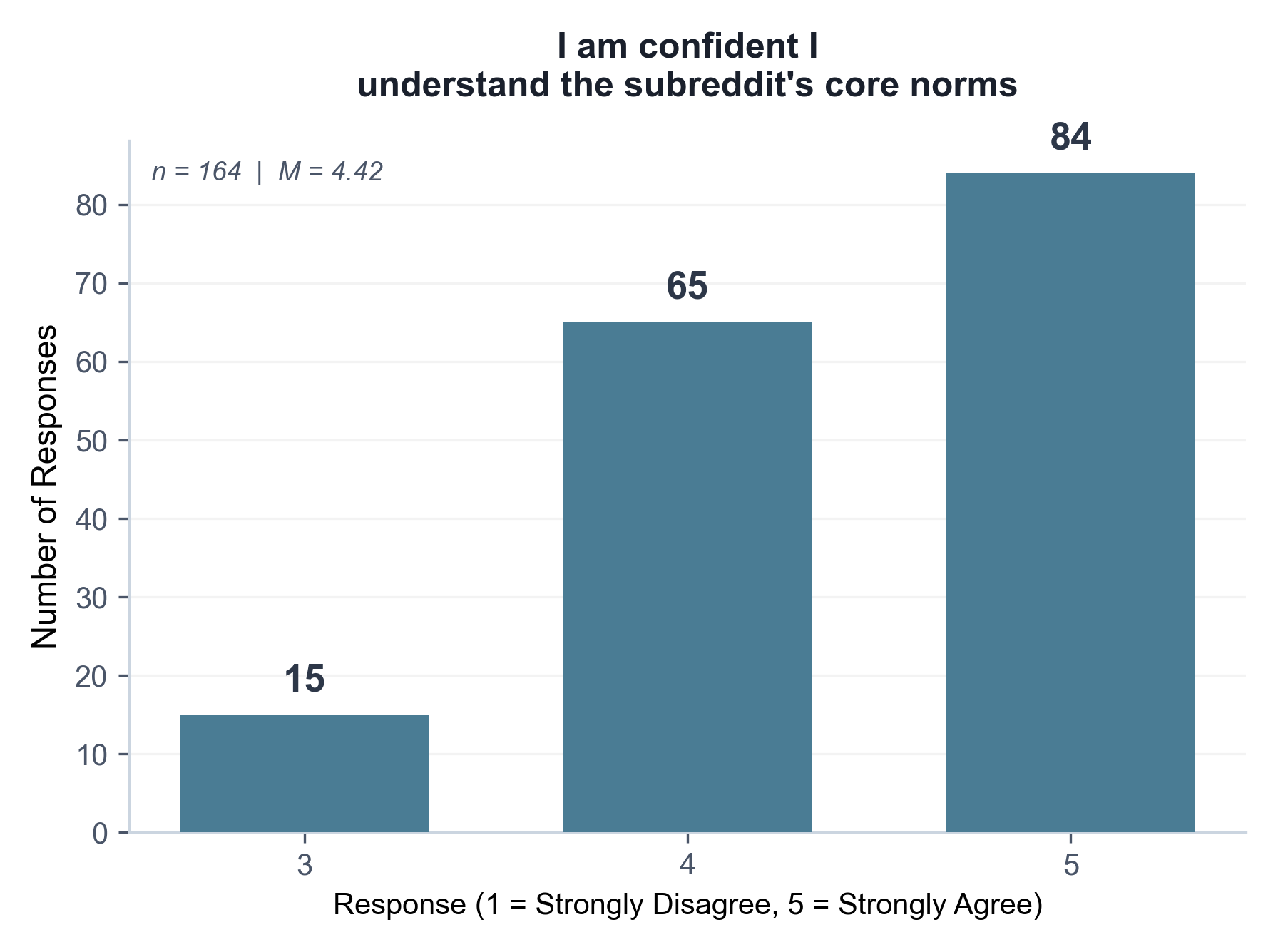}
\caption{Question 4.}
\label{fig:q4}
\end{subfigure}

\vspace{0.6em}

\begin{subfigure}[t]{0.49\linewidth}
\centering
\includegraphics[width=\linewidth]{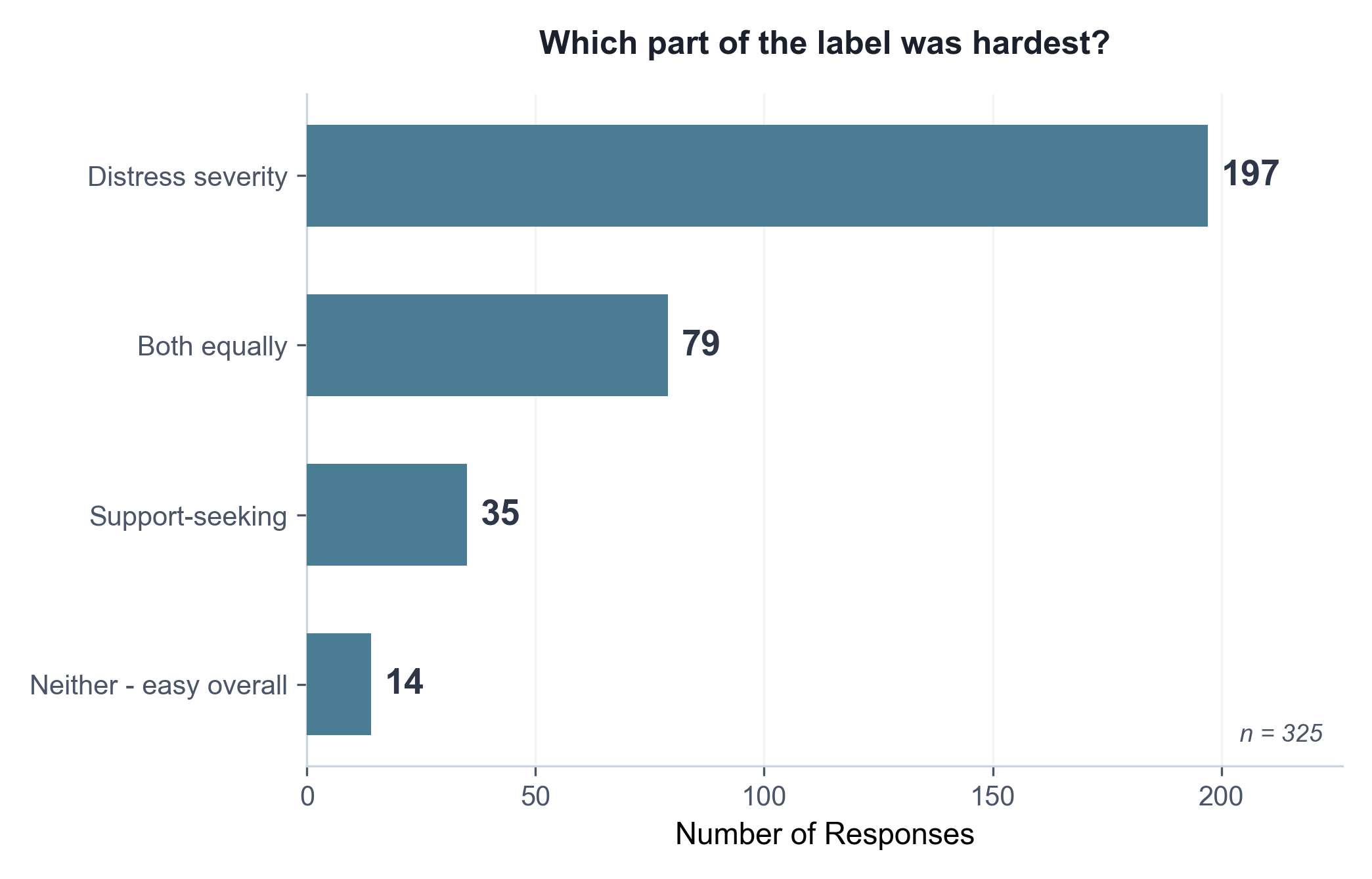}
\caption{Question 5.}
\label{fig:q5}
\end{subfigure}\hfill
\begin{subfigure}[t]{0.49\linewidth}
\centering
\includegraphics[width=\linewidth]{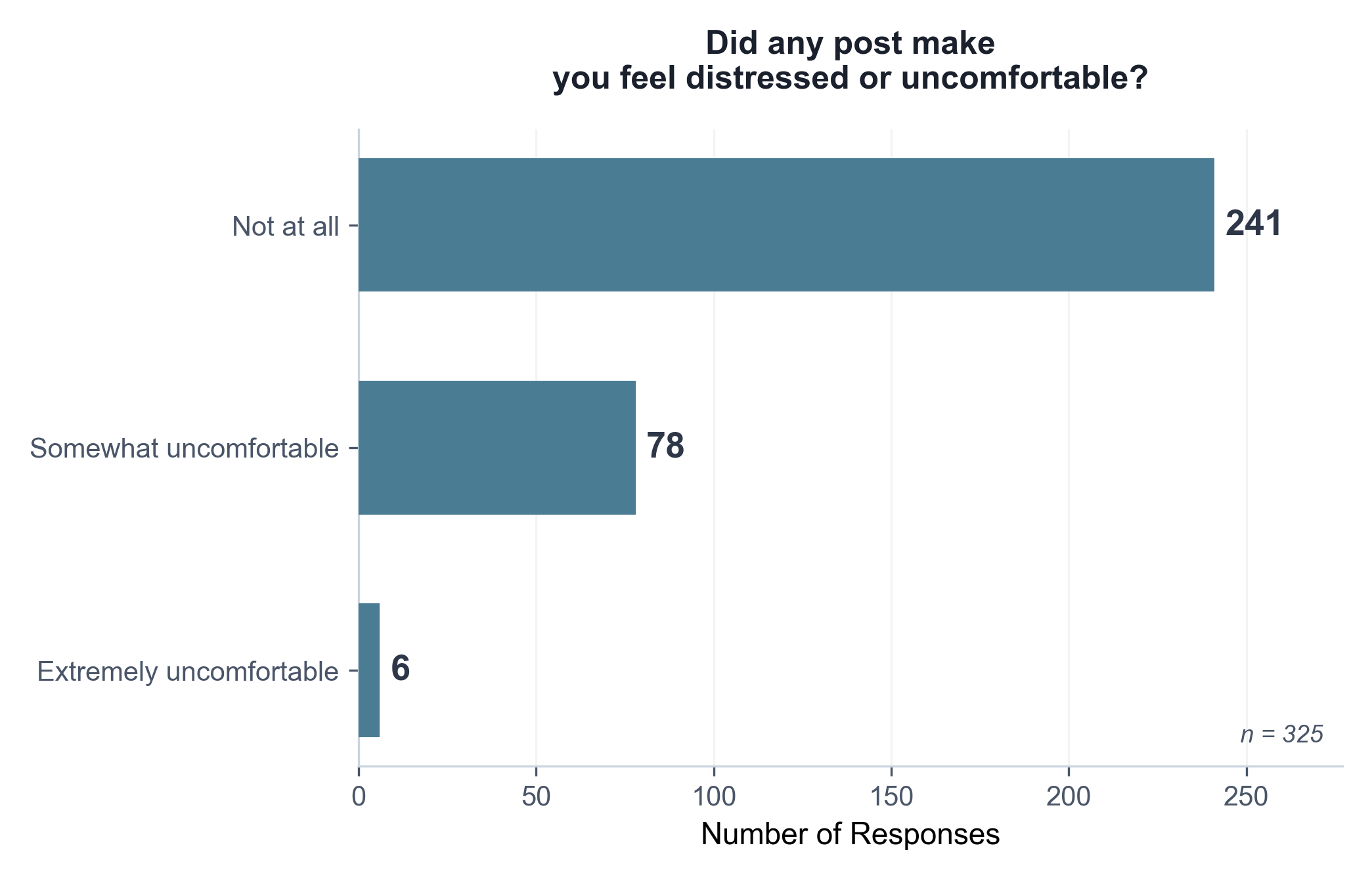}
\caption{Question 6.}
\label{fig:q6}
\end{subfigure}

\caption{Distributions for the post-task survey questions 1--6.}
\label{fig:poststudy_all}
\end{figure*}

We show the distribution of responses to the closed-ended questions in Figures~\ref{fig:q1}--\ref{fig:q6}. We qualitatively analyze participants' answers to the open-ended questions and document our findings in Section~\ref{app:qualitative}.

\section{Statistical model specification}
\label{app:models}

\paragraph{Mixed-effect models.} To test our pre-registered hypotheses, we model individual rater judgments using cross-classified mixed-effects logistic regression. Our primary outcomes are two binary variables:

\begin{itemize}
\item Distress $D_{ij}$: coded 1 if rater $i$ labeled post $j$ as ``Mild" or ``Moderate+" distress, and 0 if they selected ``None"
\item Support-seeking $S_{ij}$: coded 1 if rater $i$ indicated that the author is seeking support, and 0 otherwise
\end{itemize}

Each post belongs to one of six author communities and each rater belongs to a rater group defined by gender, parenthood status, and veteran status. Our key predictor is an indicator of IG membership, where $\mathrm{IG}_{ij}=\allowbreak \text{1}\{\text{rater } i \text{ shares the focal identity of post } j\text{’s community}\}$. For each outcome $y_{ij}\in\{D_{ij}, S_{ij}\}$, we estimate a cross-classified mixed-effects logistic model of the form:
\[ \operatorname{logit}\!\left(\Pr(y_{ij}=1)\right) = \beta_0 + \beta_{\text{group}[j]} + \gamma_{\text{group}[j]}\cdot \mathrm{IG}_{ij} + u_j + v_i,
\]
where $\beta_0$ is the overall intercept, $\beta_{\text{group}[j]}$ is a fixed effect for the subreddit of post $j$, $\gamma_{\text{group}[j]}$ is a community-specific contrast capturing how IG and OG raters differ for that community, and $u_j$ and $v_i$ are random intercepts for post $j$ and rater $i$, respectively. This specification allows the IG advantage to vary across communities, rather than assuming a single global effect. Operationally, \textbf{H1a} is evaluated by testing whether the set of $\gamma_{\text{group}}$ terms jointly differs from zero, using likelihood-ratio tests comparing the full model with a reduced model that omits the $\mathrm{IG}\times\text{community}$ interaction. We report odds ratios and 95\% confidence intervals for each community-specific contrast, separately for distress and support-seeking.

We next test whether IG raters are more likely than OG raters to agree with their own community’s aggregate label. For each post $j$ and outcome, we construct a reference label from judgments using the Dawid--Skene (DS) model\footnote{DS infers a latent label and rater-specific confusion patterns, down-weighting inconsistent annotators and yielding posterior label probabilities. This is particularly appropriate for subjective tasks where disagreement is common and rater thresholds vary systematically.}. We refer to the resulting reference labels as the IG aggregate and OG aggregate\footnote{Throughout the paper, ``aggregate'' denotes the Dawid--Skene estimated label derived from at least four IG (or four OG) ratings per post.}. To avoid circularity when the focal rater is IG, we use a leave-one-out procedure: the reference for rater $i$ is computed from the remaining IG ratings on post $j$. Let $y_j^{*}$ denote this IG aggregate label, and define an agreement indicator $a_{ij}=\text{1}\{y_{ij}=y_j^{*}\}$ which equals 1 when rater $i$’s label matches the IG aggregate for post $j$, and 0 otherwise. We fit two nested cross-classified logistic models, both sharing the same random-effects structure as above. First, a \emph{global} model,
\[
\text{logit}\,P(a_{ij}=1)=\beta_{0}+\beta_{\mathrm{group}[j]}+\gamma\cdot \mathrm{IG}_{ij}+u_{j}+v_{i},
\]
where $\gamma$ is a single pooled IG coefficient that captures the average advantage in agreement across all communities; this yields the overall OR reported in the text. Second, to recover community-specific contrasts, we replace $\gamma$ with an $\mathrm{IG}\times\text{community}$ interaction, capturing community-specific IG advantages. We also evaluate whether this is driven by participants familiar with the subreddit and test whether greater self-reported subreddit familiarity predicts higher agreement with the IG aggregate.

\paragraph{Community-specific contrasts and multiple comparisons.} Since each interaction model yields six community-specific contrasts per outcome, and we estimate these for two outcomes across two samples, the per-community tests raise a multiplicity concern. To address this, we adopt three safeguards. First, we parameterize the interaction models as $y \sim \text{community} + \text{community}\!:\!\mathrm{IG}$, so that each reported coefficient is the {\bf simple} IG--OG contrast within that community. Second, we treat the six contrasts within each outcome-by-sample combination as a family and report Holm-adjusted $p$-values ($p_{\text{holm}}$) alongside unadjusted ones. Third, we gate the per-community contrasts on the omnibus likelihood-ratio test for that family. The pooled IG coefficient $\gamma$ in the global model is the pre-registered test and is therefore reported unadjusted.

\section{Qualitative analysis of rater divergence}\label{app:qualitative}

To give more context to the quantitative patterns reported for RQ1, and in light of some surprising community-level results, we conducted qualitative analysis of both participants’ survey responses and the disputed posts.

We adopted a multi-stage, computationally assisted approach that combines content-based clustering with purposive sampling and thematic analysis. We first computed semantic embeddings for post titles and bodies using a sentence-transformer model to represent posts in latent space. We then used dimensionality reduction and unsupervised clustering method to identify broad topical groupings, enabling us to sample contested posts across diverse content types rather than within a single theme. This procedure yielded four interpretable clusters that broadly reflected (i) veterans and financial strain, (ii) gender identity and masculinity, (iii) parenting stress and everyday crises, and (iv) general relationship and social issues.

To capture IG/OG divergence at the post level, we defined two complementary indicators: (1) a measure of IG versus OG label discordance, and (2) a measure of differential alignment with the IG consensus. Within each topical cluster, we selected 15 posts which maximized both measures and whose aggregated IG and OG severity judgments differed. This resulted in a total corpus of 60 posts, ensuring our qualitative analysis focuses on high-information cases while maintaining topical breadth.

We conducted a reflexive thematic analysis of the sampled posts, first developing cluster-specific themes about how distress was expressed and interpreted, and then synthesizing these into cross-cluster themes that explained recurring sources of IG/OG divergence. We triangulated these themes with participants’ open-ended reflections to identify interpretive cues that raters explicitly referenced.

\subsection{Findings}

We identified the following distinct themes which consolidate into four major categories that capture the fundamental sources of disagreement. All excerpts drawn from Reddit have been lightly paraphrased and potentially identifying details altered to reduce the likelihood that original posts or authors can be re-identified.

\paragraph{Decoding Masked Communication.}
This theme captures how IG members recognize when community members mask or minimize their true distress through culturally-embedded communication patterns, while OG members tend to take surface-level statements literally. For example, IG members demonstrated acute awareness of indirect communication:

\IGParticipant{``In a lot of instances, people weren't directly stating their emotions. There was a lot of implied things. So, for clarification, I'm a veteran as well, and we have a habit of hiding our true feelings/emotions and never directly stating anything. A lot of what we feel has to be read between the lines.'' -- Veteran (IG)}

\IGParticipant{``There were some posts where you could tell the person was in `distress' but they didn't really outwardly express it. They didn't say things like `I'm anxious/sad/worried'. It would be a lot of `well you know I feel like maybe'.'' -- NonBinary (IG)}

On the other hand, OG members lacked this cultural fluency and struggled with indirect expression:

\Participant{``It was challenging to distinguish between a post that subtly signaled a desire for engagement and one that was purely expressive.'' -- NonBinary (OG)}

\Participant{``There were some posts that were more cryptic and evasive than others.'' -- Veterans (OG)}

A post from the r/Veterans subreddit exemplifies this pattern perfectly. The author repeatedly used disclaimers throughout: \Post{I’m not in a crisis, I just need somewhere to let it out. Please don’t feel sorry for me; I’m tough, and I’ve made it through worse.} yet goes on to say:

\Post{I went to sleep with my hands tightly clenched. I couldn’t tell if it was a heart attack coming on or a PTSD spiral.}

While IG members rated this post to exhibit Moderate+ distress, the OG rated it as Mild. On the other hand, posts using humor or light-hearted language also served as a barrier that IG members recognized (having used similar coping mechanisms themselves), but also noted the challenge of accurately decoding it due to their experience. Two IG members for r/Mommit described how uncertainty arose due to: \IGParticipant{``the tone of the author and the connection [they] felt to the stories'' -- Mommit (IG)} with another member describing how they \IGParticipant{``could tell there was more to the story, this was just a snippet.'' -- Mommit (IG)} For example, a r/Mommit post from a pregnant woman used movie references to describe their distress: \Post{I swear this little one is wedged in here tighter than\ldots{} anything wedged insanely tight\ldots{} I genuinely think I’m headed for an "Alien"-style moment some night before bed… where the baby suddenly out of nowhere tears through me and I fall apart.} This post was rated Moderate+ by IG members, while the OG did not perceive any psychological distress.

\paragraph{Community Baseline Calibration.}
IG members understand the `baseline' of their community, i.e.\ what experiences are shared, what developmental processes are expected, and what phenomena are universal, while OG members may pathologize due to lacking the necessary context. For example, experiences that are common within specific communities (fireworks triggers for veterans, period dysphoria for non-binary individuals, chronic sleep deprivation for parents) are normalized by IG members, which leads them to lower their perceived severity.

\IGParticipant{``I found that I was more confident rating the posts that included feeling I myself have felt, like anxiety.'' -- Mommit (IG)}

\IGParticipant{``Many of these problems I have had and would consider them to be fleeting since they are all far in the past.'' -- daddit (IG)}

\IGParticipant{``Sometimes I felt that what the poster was talking about might not affect me as much as it might affect them so it was difficult to find the right distress severity.'' -- TwoXChromosomes (IG)}

OG members lacked this contextual baseline, resulting in lower confidence.

\Participant{``At times I was uncertain of what the person's mental state was. Something I might consider trivial could be very distressing to another person, and because of this reason it was difficult to accurately rate some of these posts.'' -- Mommit (OG)}

A veteran describes July 4th fireworks as a shared, annual challenge: \Post{I know I’m not the only one who struggles with fireworks. And honestly, it feels like it’s gotten worse over the years, not better\ldots{} When they’re sudden and random, coming from different distances, that’s the worst part. The kind that makes me feel like I’m about to lose touch with reality.} Despite vivid symptom descriptions, the IG labeled the poster's distress as Mild, understanding that this is a community-wide, recurring experience. On the other hand, the OG rated it as Moderate+.

A similar pattern emerged for r/NonBinary posts. A non-binary author recounts how they started identifying as non-binary without having ``official'' gender dysphoria and now feel ashamed and unsure if their identity is real. IG raters assigned None for distress severity, whereas the OG marked the same post as Moderate+. While IG members perceived this post as a common pattern of struggling with labels and internalizing external invalidation as shame, OG raters lacked such baseline and instead anchored on the intensity of words like: \Post{\ldots{} always feeling ashamed constantly.} Without prior exposure to how ubiquitous these experiences are in non-binary communities, shame and confusion may be read as evidence of more serious mental health risk.

\paragraph{Contextual Integration.}
This theme captures how IG members draw on specialized knowledge and lived experience to integrate clinical markers, systemic barriers, compounding stressors, relationship dynamics, and protective factors when judging severity, while OG raters are more likely to focus on surface distress language and definitional thresholds in isolation. IG raters frequently described how pervasive symptoms seemed and how much they might interfere with daily life, even when posters did not spell this out.

\IGParticipant{``In some examples there weren't clear statements about how feelings were affecting daily life, but they did seem pervasive, so I struggled with those if they were Mild or Moderate+.'' -- NonBinary (IG)}

In contrast, OG raters repeatedly emphasized difficulty mapping posts onto category definitions. Rather than inferring impact from context, they tended to rely more on the provided codebook.

\Participant{``I was struggling with the distinction between mild and moderate.'' -- Veterans (OG)}
\Participant{``It was tricky to decide between moderate and severe. Moderate to me means someone has mild anxiety or worries. Severe means depression, trauma, body images, etc.'' -- AskMen (OG)}

IG members also showed greater sensitivity to compounding and systemic stressors. A veteran's post listing escalating financial pressures, unaffordable healthcare, childcare and housing, and a sense that

\Post{\ldots{} the world is actively working against [their] success as a father.}

was rated Moderate+ by IG raters who recognized the cumulative threat to identity and stability. OG raters, by contrast, were more likely to see the same post as Mild, treating each stressor as an isolated burden rather than a crisis.

Relationship dynamics were another site of divergence. In a r/Mommit post, a poster describes how their mother keeps kissing their baby despite being told not to, then denies it, and how this behavior echoes the poster's own childhood experiences of having their boundaries ignored and dismissed. IG raters, drawing on their parenting experiences, interpreted this as a pattern of gaslighting and bodily boundary violations and labeled it Moderate+. OG raters, however, frequently saw only a family disagreement and rated its distress level as None, missing how the present interaction reactivates earlier experiences of not being believed or protected. Overall, contextual integration reflects IG members' ability to ``see the whole picture'', while OG raters tend to evaluate only the visible slice presented in the post.

\paragraph{Temporal Anchoring.}
This theme captures a divergence in how annotators weigh current versus historical content when assessing distress severity. IG members tend to rate based on current emotional state and functioning, while OG members often anchored to historical trauma, regardless of present recovery status. For example, a post from r/Veteran described severe past trauma, but current stability: \Post{One night, I tried to end my life and was admitted to a mental health facility.\[\ldots{}\]Since then, taking my medication regularly has helped me stay stable so much so that I was able to return to my career earlier this year.} Such posts received significantly different ratings from the groups, with IG members labeling None and OG members labeling it as Moderate+. Similarly, another recovery narrative from the same subreddit:

\Post{I was totally by myself, like I was drowning. I truly didn’t believe I’d survive through another year if nothing shifted. Then I went to my nearby Vet Center, fell apart, and they’ve already hooked me up with a therapist.}

received a similar gap in labeling, with IG members labeling it as Mild and OG members labeling Moderate+. However, both groups acknowledged temporal ambiguity.

\IGParticipant{``Whenever they talk about feeling like that in the past, it made me unsure how I should rate it, if they don't currently feel like that in the present.'' -- Veteran (IG)}
\IGParticipant{``Sometimes, I'm not sure if the person is still processing either trauma or negative feelings.'' -- NonBinary (IG)}
\Participant{``Sometimes events are past tense so while the situation may have been distressing at the time, it was hard to tell if the person currently felt emotional distress.'' -- daddit (OG)}
\Participant{``Sometimes, people \ldots{} discussed having had an emotional reaction in the past but not in current terms. It was tough to have confidence in those situations.'' -- NonBinary (OG)}

\paragraph{Conclusion.} We propose these themes as the dimensions of disagreement between IG and OG annotators, namely:
\begin{itemize}
    \item \textbf{Decoding Masked Communication:} The ability to interpret indirect, minimized, humorous, or culturally-embedded expressions of distress.
    \item \textbf{Community Baseline Calibration:} Knowledge of what is ``normal'' versus concerning within a specific community baseline.
    \item \textbf{Contextual Integration:} Recognition of clinical markers, systemic factors, compounding stressors, and protective cues within the full context of a post.
    \item \textbf{Temporal Anchoring:} Whether raters prioritize present functioning or weight historical distress more heavily.
\end{itemize}

Together, these four dimensions provide a framework for understanding what shared group membership equips annotators with when assessing psychological distress.

\section{Qualitative analysis of elicited LLM rationales}\label{app:trace-analysis}

To complement the quantitative error analysis in RQ2b, we conducted a thematic analysis of the chain-of-thought rationales elicited from reasoning LLMs, aiming to characterize the strategies models \textit{articulate} when assessing distress and how these shift under conditioning. This section details the rationale patterns we identified and illustrative excerpts. We treat such traces as \textit{diagnostic evidence} since chain-of-thought traces are not guaranteed faithful records of the computations that determine the final label \cite{turpin2023language}.

\subsection{Post sampling and coding procedure}

We selected 37 posts for chain-of-thought analysis by cross-referencing model predictions (Olmo-3-7B-Think and Qwen3-30B-A3B under $\varnothing$ and $CE$ prompting) against IG aggregate labels. Table~\ref{tab:trace-categories} outlines the five categories that cover the landscape of conditioning effects. Within each category, we maximize subreddit and IG-severity diversity. For all 37 posts, we extracted and analyzed reasoning traces from both models under both conditions, yielding 148 model-condition trace pairs. We coded these in two passes: open coding to surface recurring reasoning moves, followed by systematic application of a codebook capturing the five rationale patterns described below.

\begin{table*}[t]
\centering
\setlength{\tabcolsep}{10pt}
\renewcommand{\arraystretch}{1.3}
\begin{tabular}{@{}l c c c c l@{}}
\toprule
\textbf{Category} & \textbf{Posts} & 
\textbf{$\varnothing$ vs.\ IG} & \textbf{$CE$ vs.\ IG} & \textbf{Requires IG $\neq$ OG} &
\textbf{Diagnostic purpose} \\
\midrule
Conditioning fixes      & 9 & $\uparrow$  & $=$          &              & What conditioning corrects \\
Overcorrection          & 9 & $\uparrow$  & $\downarrow$ &              & Where conditioning overshoots \\
Persistent over-est.    & 7 & $\uparrow$  & $\uparrow$   &              & What resists correction \\
Correct on hard         & 6 &             & $=$          & $\checkmark$ & Navigating ambiguity \\
Under-estimation        & 6 &             & $\downarrow$ &              & False negatives from conditioning \\
\bottomrule
\end{tabular}
\caption{Diagnostic categories used for trace sampling. Arrows show the direction of model prediction relative to the IG aggregate: $\uparrow$ over-estimates, $=$ matches, $\downarrow$ under-estimates. Empty cells indicate the criterion is not constrained for that category. Conditions are evaluated under vanilla ($\varnothing$) and contextualized-with-examples ($CE$) prompting.}
\label{tab:trace-categories}
\end{table*}

\subsection{Findings}

We identified five recurring patterns in the elicited rationales.

\paragraph{Salient-phrase anchoring.} The trace centers its severity reasoning on a small set of emotionally loaded words or metaphors (``existential crisis,'' ``lost all hope,'' ``feeling like a fake,'' ``terrified''), with less weight given to whether the post describes concrete impairment, coping, or temporal resolution. Clinical-sounding or hyperbolic self-labels are treated as direct evidence for rubric thresholds (e.g., mapping ``lost all hope'' to the ``explicit hopelessness'' criterion for Moderate+), even when the broader narrative context suggests a bounded vent or colloquial register. For example, an r/AskMen poster asks how other men with high incomes but no family feel, writing: \Post{I feel like a mutilated victory. Honestly, I lost all hope.} Both models under both conditions assigned Moderate+ with high confidence. Qwen3's vanilla trace reasons: \Trace{The author explicitly states `I lost all hope,' which constitutes explicit hopelessness, a key criterion for Moderate+ severity. This is not a fleeting or trivial emotion but a profound expression of despair\ldots{} No mitigating factors (e.g., references to seeking help, hopeful outlook) are present.} The conditioned trace follows identical logic: \Trace{The phrase `lost all hope' is a well-established clinical indicator\ldots{} The criteria for Moderate+ includes `explicit hopelessness,' and the author's statement is a clear, unambiguous expression of this.} Despite the short post containing no evidence of functional impairment and the community framing of a peer poll, the single phrase locks both conditions into Moderate+.

\paragraph{Impairment over-inference.} The trace infers rubric-style ``interference with daily life'' from domain cues that are vague, ambiguous, or primarily situational (e.g., mapping job-search frustration to ``occupational impairment'' or childcare exhaustion to ``daily-life interference'') without explicit textual evidence that the poster's psychological state is disrupting their functioning. For example, an r/AskMen poster describes an ``existential crisis'' because partners say they do not feel needed in her life, which IG raters labeled Mild. In the vanilla condition, Qwen3 maps ``existential crisis'' to ``worthlessness'' and ``don't feel needed'' to relational impairment, concluding Moderate+:

\Trace{The existential crisis term further elevates severity beyond Mild, as it indicates profound, ongoing psychological distress rather than transient anxiety.}

In the conditioned trace, the same model downgrades to Mild by reframing the phrase:

\Trace{The term `existential crisis' is used descriptively but not as a severe symptom\ldots{} The author's focus on seeking solutions (`how can your girlfriend make you feel needed?') reflects the coping behavior required for Mild, unlike Moderate+ examples that show clear functional impairment.}

Here, exemplar-driven normalization (comparing to a similar in-prompt post labeled Mild) and stricter rubric enforcement (demanding functional evidence) corrected the vanilla error.

\paragraph{Rubric literalism.} The trace applies strict checklist logic, treating only named affective states (anxiety, sadness, depression) as valid evidence for distress. Frustration, anger, or pain-adjacent venting are excluded because they are ``not in the rubric list'', and physical symptoms or bureaucratic frustrations are dismissed without explicit affect labels. For example, a veteran posts about guilt over an impending 100\% disability rating despite never having seen combat, describing ``icky thoughts'' about self-harm and feeling like a scam. Such post was rated Mild by  the IG. Qwen3's vanilla trace assigns Moderate+ by reading ``icky thoughts'' as euphemistic suicidal ideation. The conditioned trace, however, maps the post onto a procedural in-prompt exemplar:

\Trace{The community consistently labels posts expressing general guilt or self-doubt about benefits (without functional impairment) as `None,' as seen in POST4\ldots{} The distress is about perceived unworthiness, not active symptom interference.}

The model assigns None with high confidence, undershooting the IG Mild consensus by treating a post with self-harm language as a routine benefits question.

\paragraph{Temporal and resolution discounting.} The trace's decisive reasoning move concerns \textit{when} distress is situated in the narrative. Past-tense or resolved distress (celebratory endings, reflective tone) is discounted to None, while embedded backstories of worry are treated as ongoing. This parallels the \textit{Temporal Anchoring} theme from our human-rater qualitative analysis, but traces apply it more rigidly: a post describing weeks of anxiety followed by a positive outcome may receive None despite IG raters recognizing that the distress warranted Mild. For instance, an r/NonBinary poster shares a celebratory coming-out post that opens with \Post{I've been worried about how this would go down for weeks, would I find the right words? Is it better just to pretend I'm not NB? Am I throwing my life away?} before closing with relief: \Post{All in all it went well and feels so much better to be truthful once again.} IG raters assigned Mild, crediting the explicitly named weeks of worry and somatic cue (poor sleep) as genuine first-person distress. Qwen3's vanilla trace states:

\Trace{The author describes past distress \ldots{} but explicitly states this has been resolved with a positive outcome. The distress was temporary and resolved, fitting the `None' category per criteria (no current negative state).}

The conditioned trace follows the same logic while additionally invoking a community exemplar:

\Trace{While they mention past worries, these are framed as resolved past concerns, not current distress \ldots{} This is consistent with Example 4 (LABEL4: None), where a post expressing past shame was classified as `None' because the distress was not current or impairing.}

The post's celebratory present tense fully overrides weeks of named anxiety across both conditions.

\paragraph{Exemplar-driven normalization.} In conditioned runs, the trace invokes in-context exemplars or ``how this subreddit reads vents'' framing to reclassify posts as typical, fleeting, or situational. The trace maps the target post onto the closest exemplar by genre (e.g., ``this is a POST4-style procedural question'') and adopts that exemplar's label. This is the pattern most consistently associated with conditioning-related reductions in over-estimation, but it also appears in over-correction cases when the exemplar match is superficial (aligning with the aforementioned example of discounting a veteran's guilt about disability benefits as ``just a system question'' because it superficially resembles a procedural \hbox{exemplar}).

\subsection{Conclusion}

The five rationale patterns provide a diagnostic taxonomy of how LLMs articulate their distress judgments. Salient-phrase anchoring and impairment over-inference are the dominant patterns in over-estimation traces, while rubric literalism and exemplar-driven normalization are the primary patterns associated with conditioning-related corrections, but both also appear in over-correction cases when applied too rigidly. Temporal discounting co-occurs with errors in either direction depending on the post's narrative structure. These patterns parallel the human-rater disagreement themes from Appendix~\ref{app:qualitative}: exemplar-driven normalization parallels \textit{Community Baseline Calibration}, temporal discounting maps onto \textit{Temporal Anchoring}, and the failure modes of salient-phrase anchoring echo a breakdown of \textit{Contextual Integration}, suggesting that the gap between model and in-group judgments are not random noise but a structured deficit in the same interpretive capacities that distinguish in-group from out-group human raters.

\section{LLM prompts and generation examples}\label{app:prompts}

Figures~\ref{fig:v_prompt}, \ref{fig:c_prompt}, and~\ref{fig:ce_prompt} detail the prompts we use for the LLM generations. Identical prompts were used for the classification of support-seeking. Across all models, we generated responses with temperature = 0.7, allowing up to 8192 newly generated tokens. Figure~\ref{fig:e_gen_example} outlines a generation example. Model outputs are parsed into the same label space as the human judgments.

\begin{figure*}[!htb]
\centering
\begin{minipage}{0.98\linewidth}
\begin{mdframed}[
linewidth=0.8pt,
linecolor=black,
innerleftmargin=8pt,
innerrightmargin=8pt,
innertopmargin=8pt,
innerbottommargin=8pt
]
\small
\setlength{\parindent}{0pt}

Instruction: You are an exceptional assistant whose task is to carefully read social media posts and assess their psychological / emotional distress severity level. You are to abide by the following instructions, provide reasoning for your classification, and reflect on your confidence level.

\par\medskip
---

\par\medskip
\# Response Format

You MUST end your response with EXACTLY this format - no variations allowed:

First, provide your reasoning in these sections:
\par
- Severity reasoning\\
- Confidence reasoning

\par\medskip
Then END with this EXACT line:

Final answer: \{``severity": ``\textless value\textgreater", ``confidence": ``\textless value\textgreater"\}

\par\medskip
Only possible values for severity:
\par
- None\\
- Mild\\
- Moderate+

\par\medskip
Only possible values for confidence:
\par
- High\\
- Low

\par\medskip
Remember, you must end your response with a JSON.

\par\medskip
---

\par\medskip
\# Instructions

\{\{ codebook \}\}

\par\medskip
---

\par\medskip
\# Post

The post you are to classify is as follows:

Title: \{\{ post\_title \}\}

Post: \{\{ post\_text \}\}

\par\medskip
---
\end{mdframed}
\end{minipage}

\caption{LLM Prompt for the vanilla condition ($\varnothing$) for distress severity classification.}
\label{fig:v_prompt}
\end{figure*}

\begin{figure*}[!htb]
\centering
\begin{minipage}{0.98\linewidth}
\begin{mdframed}[
linewidth=0.8pt,
linecolor=black,
innerleftmargin=8pt,
innerrightmargin=8pt,
innertopmargin=8pt,
innerbottommargin=8pt
]
\small
\setlength{\parindent}{0pt}

Instruction: You are a community member from the subreddit \/\{\{ subreddit \}\} whose task is to carefully read social media posts from this subreddit and assess their psychological / emotional distress severity level. You are to abide by the following instructions, provide reasoning for your classification, and reflect on your confidence level. Whenever you are unsure, think about how members of the assigned community would generally interpret such posts.

\ldots
\end{mdframed}
\end{minipage}

\caption{LLM Prompt for the contextualized condition ($C$) for distress severity classification. The remaining prompt is identical as the $\varnothing$ condition.}
\label{fig:c_prompt}
\end{figure*}

\begin{figure*}[!htb]
\centering
\begin{minipage}{0.98\linewidth}
\begin{mdframed}[
linewidth=0.8pt,
linecolor=black,
innerleftmargin=8pt,
innerrightmargin=8pt,
innertopmargin=8pt,
innerbottommargin=8pt
]
\small
\setlength{\parindent}{0pt}

Instruction: You are a community member from the subreddit \/\{\{ subreddit \}\} whose task is to carefully read social media posts from this subreddit and assess their psychological / emotional distress severity level. You are to abide by the following instructions, provide reasoning for your classification, and reflect on your confidence level. Whenever you are unsure, think about how members of the assigned community would generally interpret such posts. To help in making your decision, you are provided with example posts annotated by other community members.

\ldots

\# Example Posts Annotated by Other Community Members

\{\{ example\_posts \}\}

\ldots
\end{mdframed}
\end{minipage}

\caption{LLM Prompt for the contextualized with examples condition ($CE$) for distress severity classification. The remaining prompt is identical as the $\varnothing$ condition.}
\label{fig:ce_prompt}
\end{figure*}

\begin{figure*}[!htb]
\centering
\begin{minipage}{0.98\linewidth}
\begin{mdframed}[
linewidth=0.8pt,
linecolor=black,
innerleftmargin=8pt,
innerrightmargin=8pt,
innertopmargin=8pt,
innerbottommargin=8pt
]
\small
\setlength{\parindent}{0pt}
$<$think$>$

Okay, let's tackle this classification step by step. The user provided a post and wants to know the severity of emotional or psychological distress. The instructions are clear, so I need to follow them carefully. First, I'll read through the post again to understand what's being expressed. The post details an incident where the person missed a phone call related to their HLR process. They explain that the call came in a few days after filing, but they didn't answer because it was an unknown number. After some research, they found out it was the HLR call, so they called back. The main points are frustration over missing the call, some confusion about the timing (calls coming in 2 months +/- 2 weeks), and a cautionary note to others about answering unknown numbers.

Now, looking for the key elements of severity. The criteria for None, Mild, and Moderate+ need to be checked.

Starting with None: There's no first-person expression of a negative internal state like anxiety, sadness, depression, etc. The post is more about a logistical issue and a warning to others. The author mentions frustration (``Ugh") but that's more of an exclamation than a deep emotional state. There's no mention of feeling anxious, sad, depressed, or any significant internal distress. The main emotion here seems to be frustration or annoyance, which might be mild but not crossing into a moderate level.

Looking at Mild criteria: Explicit mention of a negative internal state but no functional impairment, or coping/optimism present. The person did express some frustration, but does that count as a negative internal state? The instructions say ``explicitly mentioned negative internal state like anxiety, sadness, depression, panic, etc." ``Ugh" might be mild annoyance, but maybe not enough to qualify as a recognized negative internal state. Also, the post ends with a helpful note to others, showing some constructive intent. They didn't dwell on their own distress but instead provided advice. That could indicate some coping or optimism.

Moderate+ would require interference with daily life, hopelessness, worthlessness, or self-harm thoughts. The post doesn't mention any issues affecting work, sleep, relationships, or anything like that. The problem was about a missed call and some administrative process. There's no indication of the distress causing functional impairment. Also, no signs of hopelessness or suicidal thoughts.

So, the main points against Moderate+ are the absence of functional impairment and the lack of severe emotional states. The frustration is there, but it's more of a temporary inconvenience rather than a deep psychological issue. The fact that they're trying to help others by posting the warning might suggest they're managing their feelings, which aligns with coping or mild distress without severe impact.

Confidence-wise, I'm a bit unsure if ``Ugh" is enough for a Mild severity. The criteria for Mild requires an explicit mention of a negative internal state (like anxiety, sadness, etc.). The word ``Ugh" is more of an exclamation than a declaration of an emotional state. For example, ``I feel anxious" would be explicit, but ``Ugh" is more like a complaint. The example given for Mild is ``I'm a bit anxious about finals..." which includes the emotional state. Since the post doesn't use such explicit language, maybe it's not meeting the Mild criteria. Then the next step down is None.

Wait, but the post does mention missing the call and the frustration. Could that frustration be considered a mild negative internal state? The instructions for None say that if the distress is about someone else, hypothetical, fleeting, or trivial, it's None. Since this is the author's own experience, it's not about someone else. The frustration is their own, but is it fleeting? They seem to have resolved it by calling back and giving a heads-up. The duration isn't long-term. So maybe it's None because there's no explicit mention of a recognized mental health issue, just a temporary inconvenience.

\ldots

$</$think$>$

Severity reasoning: The post describes a minor frustration about missing a phone call related to an administrative process but does not explicitly express a recognized negative internal state (e.g., anxiety, sadness, depression). The tone is practical and solution-oriented, with no mention of functional impairment, hopelessness, or severe emotional distress. The frustration is situational and transient, lacking the specificity or emotional weight required for higher severity.

Confidence reasoning: The classification hinges on the absence of explicit emotional language (e.g., ``I feel anxious" or ``I am depressed") and the focus on a logistical issue resolved through action. While mild annoyance is implied, it does not meet the criteria for ``Mild" due to the lack of explicit internal state declaration or functional impact. Confidence is low because the line between frustration and trivial concern is ambiguous without clearer emotional cues.

Final answer: \{ ``severity": ``None", ``confidence": ``Low" \}

\end{mdframed}
\end{minipage}

\caption{Illustrative generation showing a model’s reasoning (specifically, Olmo-3-7B-Think) for distress classification.}
\label{fig:e_gen_example}
\end{figure*}

\section{Inter-annotator reliability}\label{app:agreement}

Table~\ref{tab:iar-by-group} reports within-group inter-annotator reliability (Krippendorff’s $\alpha$) for distress severity and support-seeking labels, computed separately for IG and OG raters across communities. As can be noted, agreement is generally moderate for severity (approximately $\alpha \approx .29$–$.54$) and moderate for support-seeking (approximately $\alpha \approx .34$–$.50$), with some variation by subreddit.

\begin{table*}
\centering
\small
\setlength{\tabcolsep}{6pt}
\renewcommand{\arraystretch}{1.15}

\begin{tabular*}{\textwidth}{@{\extracolsep{\fill}}l c c c c c c@{}}
\toprule
& \multicolumn{3}{c}{\textbf{In-group (IG)}} & \multicolumn{3}{c}{\textbf{Out-group (OG)}} \\
\cmidrule(lr){2-4} \cmidrule(lr){5-7}
\textbf{Community} &
$N$ ratings & $\alpha$ (Sev.) & $\alpha$ (Seek.) &
$N$ ratings & $\alpha$ (Sev.) & $\alpha$ (Seek.) \\
\midrule
r/AskMen & 800 & .440 & .343 & 800 & .488 & .350 \\
r/daddit & 800 & .374 & .420 & 800 & .439 & .448 \\
r/Mommit & 800 & .350 & .490 & 800 & .442 & .441 \\
r/NonBinary & 796 & .416 & .475 & 798 & .317 & .502 \\
r/TwoXChromosomes & 796 & .293 & .474 & 796 & .340 & .477 \\
r/Veterans & 801 & .530 & .363 & 800 & .540 & .379 \\
\bottomrule
\end{tabular*}

\caption{Inter-annotator reliability (Krippendorff’s $\alpha$) within IG and OG rater groups for distress severity (Sev.) and support-seeking (Seek.) labels, reported by community.}
\label{tab:iar-by-group}
\end{table*}

\section{Familiarity decomposition within the IG condition}
\label{app:familiarity}

\paragraph{Motivation and design.} The IG condition combines two accounts that could each contribute to the observed agreement advantage: shared demographic identity and access to subreddit-source information. Since OG raters had neither, the current design cannot isolate the effect of identity from the effect of source disclosure. However, we can still test whether the IG advantage depends on deep prior familiarity with the subreddit. To that end, we decompose the IG condition by self-reported subreddit familiarity into two subgroups: \textbf{IG\textsubscript{NEW}} ($n = 106$ raters), who were not prior subreddit members and received brief familiarization, and \textbf{IG\textsubscript{FAMILIAR}} ($n = 57$ raters), who were already occasional or active users. Both subgroups share the focal demographic identity and subreddit disclosure, differing primarily in depth of prior community exposure. If deep familiarity were the primary driver of the IG advantage, we would expect a pronounced gradient: OG $\ll$ IG\textsubscript{NEW} $\ll$ IG\textsubscript{FAMILIAR}.

The data show a different pattern. Agreement rates increase from OG (80.5\%) to IG\textsubscript{NEW} (82.1\%) to IG\textsubscript{FAMILIAR} (82.6\%), but most of the gain occurs at the OG~$\to$~IG\textsubscript{NEW} transition. In a mixed-effects logistic model controlling for subreddit with random intercepts for post and rater, both IG\textsubscript{NEW} and IG\textsubscript{FAMILIAR} show marginal advantages over OG (OR~$= 1.16$, $p = .097$ and OR~$= 1.22$, $p = .075$, respectively), but critically, the two IG subgroups do not differ from each other (OR~$= 1.05$, $p = .674$), and the three-level familiarity factor does not improve model fit over the binary IG indicator (LRT $\chi^2(1) = 0.18$, $p = .674$). The effect is clearest in parenting communities, where both subgroups significantly outperform OG. Restricting to high-confidence annotations strengthens both effects to conventional significance ($p = .023$ and $p = .018$).

The similar performance of IG\textsubscript{NEW} and IG\textsubscript{FAMILIAR} raters suggests that the IG advantage does not depend on long-term prior familiarity with the subreddit. Even brief familiarization appears sufficient for IG raters to perform comparably to those with deeper community experience. However, because all IG raters shared the focal identity and received subreddit information, this analysis cannot tell whether the advantage comes more from identity match or from source disclosure. The main takeaway is: the IG advantage is unlikely to be driven primarily by depth of prior subreddit familiarity.

\section{Random-stratum robustness}
\label{app:random-only}

As detailed in the main text, our purposeful sampling strategy produces a targeted distribution of posts with salient distress and support-seeking cues. This targeted selection improves yield for studying distress judgments but raises the concern that IG--OG differences may be compressed and that LLM over-estimation may be inflated. The random stratum, comprising 40 posts per subreddit drawn \textit{before} any weak supervision or LLM filtering, provides a built-in validity check. This subset comprises of 1,896 annotations from 309 participants. The IG consensus distribution for severity was: None 51.9\%, Mild 28.7\%, and Moderate+ 19.4\%, substantially shifted toward lower distress compared with the full purposeful sample, confirming that the targeted selection inflates distress prevalence. Among posts with any distress, 40.1\% were rated as seeking support.

\paragraph{Results.} The random stratum revealed stronger evidence of community-specific differences than the full sample, consistent with the interpretation that the targeted distribution compresses IG--OG gaps. All contrasts below are simple IG--OG effects within the named community, with Holm adjustment applied across the six contrasts in each family; Table~\ref{tab:random-contrasts} reports the full set.

For distress detection, the subreddit $\times$ IG interaction was highly significant, $\chi^2(5) = 18.07$, $p = .003$, driven by r/daddit ($\mathrm{OR} = 7.34$, 95\% CI $[2.43, 22.23]$, $p = .0004$, $p_{\text{holm}} = .003$), where IG raters had roughly seven times the odds of detecting distress than OG raters. For support-seeking, the interaction was also significant, $\chi^2(5) = 13.68$, $p = .018$, with the largest contrast in r/Veterans ($\mathrm{OR} = 0.06$, 95\% CI $[0.006, 0.49]$, $p = .009$, $p_{\text{holm}} = .055$), where OG raters detected support-seeking more frequently than IG raters (86.3\% vs.\ 62.7\%).

Consistent with the full analysis, IG raters showed significantly higher agreement with their community's distress aggregate ($\beta = 0.28$, 95\% CI $[0.01, 0.54]$, $p = .042$; $\mathrm{OR} = 1.32$, 95\% CI $[1.01, 1.72]$), but no IG advantage emerged for support-seeking agreement ($p = .593$). The omnibus interaction for distress agreement was strongly significant, $\chi^2(5) = 21.23$, $p = .0007$, and here the dominant contrast is r/NonBinary ($\mathrm{OR} = 2.96$, 95\% CI $[1.66, 5.29]$, $p = .0002$, $p_{\text{holm}} = .001$), followed by r/daddit ($\mathrm{OR} = 1.93$, $p = .024$, $p_{\text{holm}} = .119$). For support-seeking agreement the omnibus test was not significant, $\chi^2(5) = 11.02$, $p = .051$.

\begin{table}[!htb]
\centering
\small
\setlength{\tabcolsep}{4pt}
\renewcommand{\arraystretch}{1.15}
\begin{tabular}{@{}l c c c c@{}}
\toprule
\textbf{Community} & \textbf{OR} & \textbf{95\% CI} & \textbf{$p$} & \textbf{$p_{\text{holm}}$} \\
\midrule
\multicolumn{5}{@{}l}{\textit{Distress detection} \quad $\chi^2(5)=18.07$, $p=.003$} \\
\quad r/AskMen          & 1.02 & [0.33, 3.16]  & .975  & 1.000 \\
\quad r/daddit          & 7.34 & [2.43, 22.23] & .0004 & \textbf{.003} \\
\quad r/Mommit          & 2.67 & [0.90, 7.92]  & .077  & .309 \\
\quad r/NonBinary       & 0.37 & [0.14, 1.01]  & .052  & .262 \\
\quad r/TwoXChrom.      & 0.83 & [0.27, 2.55]  & .744  & 1.000 \\
\quad r/Veterans        & 1.02 & [0.35, 3.02]  & .968  & 1.000 \\
\midrule
\multicolumn{5}{@{}l}{\textit{Support-seeking detection} \quad $\chi^2(5)=13.68$, $p=.018$} \\
\quad r/AskMen          & 3.53 & [0.20, 61.84] & .388  & 1.000 \\
\quad r/daddit          & 3.45 & [0.66, 17.90] & .141  & .705 \\
\quad r/Mommit          & 2.76 & [0.45, 16.76] & .271  & 1.000 \\
\quad r/NonBinary       & 0.73 & [0.17, 3.19]  & .678  & 1.000 \\
\quad r/TwoXChrom.      & 0.60 & [0.18, 2.08]  & .424  & 1.000 \\
\quad r/Veterans        & 0.06 & [0.01, 0.49]  & .009  & .055 \\
\midrule
\multicolumn{5}{@{}l}{\textit{Distress agreement} \quad $\chi^2(5)=21.23$, $p=.0007$} \\
\quad r/AskMen          & 0.70 & [0.33, 1.49]  & .356  & .480 \\
\quad r/daddit          & 1.93 & [1.09, 3.42]  & .024  & .119 \\
\quad r/Mommit          & 1.74 & [0.97, 3.11]  & .064  & .256 \\
\quad r/NonBinary       & 2.96 & [1.66, 5.29]  & .0002 & \textbf{.001} \\
\quad r/TwoXChrom.      & 0.66 & [0.34, 1.30]  & .229  & .480 \\
\quad r/Veterans        & 0.61 & [0.31, 1.22]  & .160  & .480 \\
\bottomrule
\end{tabular}
\caption{Random-stratum community contrasts. Each row is the simple IG--OG odds ratio within that community. $p_{\text{holm}}$ is Holm-adjusted across the six contrasts in its family. The support-seeking agreement family is omitted because its omnibus test was not significant.}
\label{tab:random-contrasts}
\end{table}

These results suggest that the modest IG alignment effects observed in the main analysis are robust to sampling strategy and, if anything, conservative: effects are larger when evaluated at natural base rates. They also show that the community carrying the effect is not stable across samples, which is why we rely on the pooled test and the omnibus heterogeneity tests for our substantive claims.

\section{LLM--IG alignment for support-seeking}
\label{app:llm-ig-seeking}

\paragraph{Evaluation.} Support-seeking is evaluated on posts where the IG consensus for distress is at least ``Mild''. Posts with no perceived distress lack ground truth for support-seeking intent, as the construct is only meaningful when distress is present. After applying this filter, 923 posts were evaluated for support-seeking classification. We caution against drawing strong conclusions from the support-seeking results given this limited coverage.

\paragraph{Results.} Table~\ref{tab:seeking-$F_1$} presents $F_1$-scores for support-seeking classification, comparing human rater performance and LLM alignment with the IG consensus. The IG-rater median (median IG human-human $F_1$) was 0.8, indicating moderate inter-rater agreement on support-seeking labels. Out-group aggregate performance exceeded this reference point ($F_1$ = 0.848). Among the evaluated LLMs, Qwen3-30B-A3B-Thinking achieved the highest performance ($F_1$ = 0.831, +3.9\%). Both Qwen3-30B-A3B-Instruct and Olmo-3-7B (both variants) under-performed relative to the human benchmark, with $F_1$-scores ranging from 0.712--0.728 (-9.0\% to -11.0\% vs.\ IG-rater median).

\begin{table}[!htb]
\centering
\small
\setlength{\tabcolsep}{6pt}
\renewcommand{\arraystretch}{1.2}

\begin{tabular}{@{}l c c@{}}
\toprule
\textbf{Group/Variant} & \textbf{$F_1$ [95\% CI]} & \textbf{vs. IG median} \\
\midrule
\textit{Human Raters} & & \\
\quad In-group (median) & 0.800 & --- \\
\quad Out-group aggregate & 0.848 [0.824, 0.870] & +6.0\% \\
\midrule
\textit{Qwen3-30B-A3B-Thinking} & & \\
\quad Vanilla ($\varnothing$) & 0.831 [0.806, 0.854] & +3.9\% \\
\quad Context ($C$) & 0.800 [0.774, 0.825] & +0.0\% \\
\quad Context + Examples ($CE$) & 0.788 [0.761, 0.812] & -1.5\% \\
\midrule
\textit{Qwen3-30B-A3B-Instruct} & & \\
\quad Vanilla ($\varnothing$) & 0.728 [0.701, 0.754] & -9.0\% \\
\quad Context ($C$) & 0.723 [0.697, 0.748] & -9.6\% \\
\quad Context + Examples ($CE$) & 0.722 [0.694, 0.747] & -9.8\% \\
\midrule
\textit{Olmo-3-7B-Think} & & \\
\quad Vanilla ($\varnothing$) & 0.725 [0.699, 0.749] & -9.4\% \\
\quad Context ($C$) & 0.721 [0.693, 0.746] & -9.9\% \\
\quad Context + Examples ($CE$) & 0.726 [0.701, 0.751] & -9.3\% \\
\midrule
\textit{Olmo-3-7B-Instruct} & & \\
\quad Vanilla ($\varnothing$) & 0.712 [0.684, 0.737] & -11.0\% \\
\quad Context ($C$) & 0.719 [0.692, 0.743] & -10.2\% \\
\quad Context + Examples ($CE$) & 0.720 [0.693, 0.744] & -10.1\% \\
\bottomrule
\end{tabular}

\caption{$F_1$-scores for support-seeking detection. The IG-rater median (0.8) represents the median pairwise $F_1$ between in-group raters; it is a reference point, not a performance ceiling.}
\label{tab:seeking-$F_1$}
\end{table}

\section{LLM conditional errors for distress}
\label{app:conditional}

\begin{figure}[!htbp]
\centering
\includegraphics[width=\linewidth]{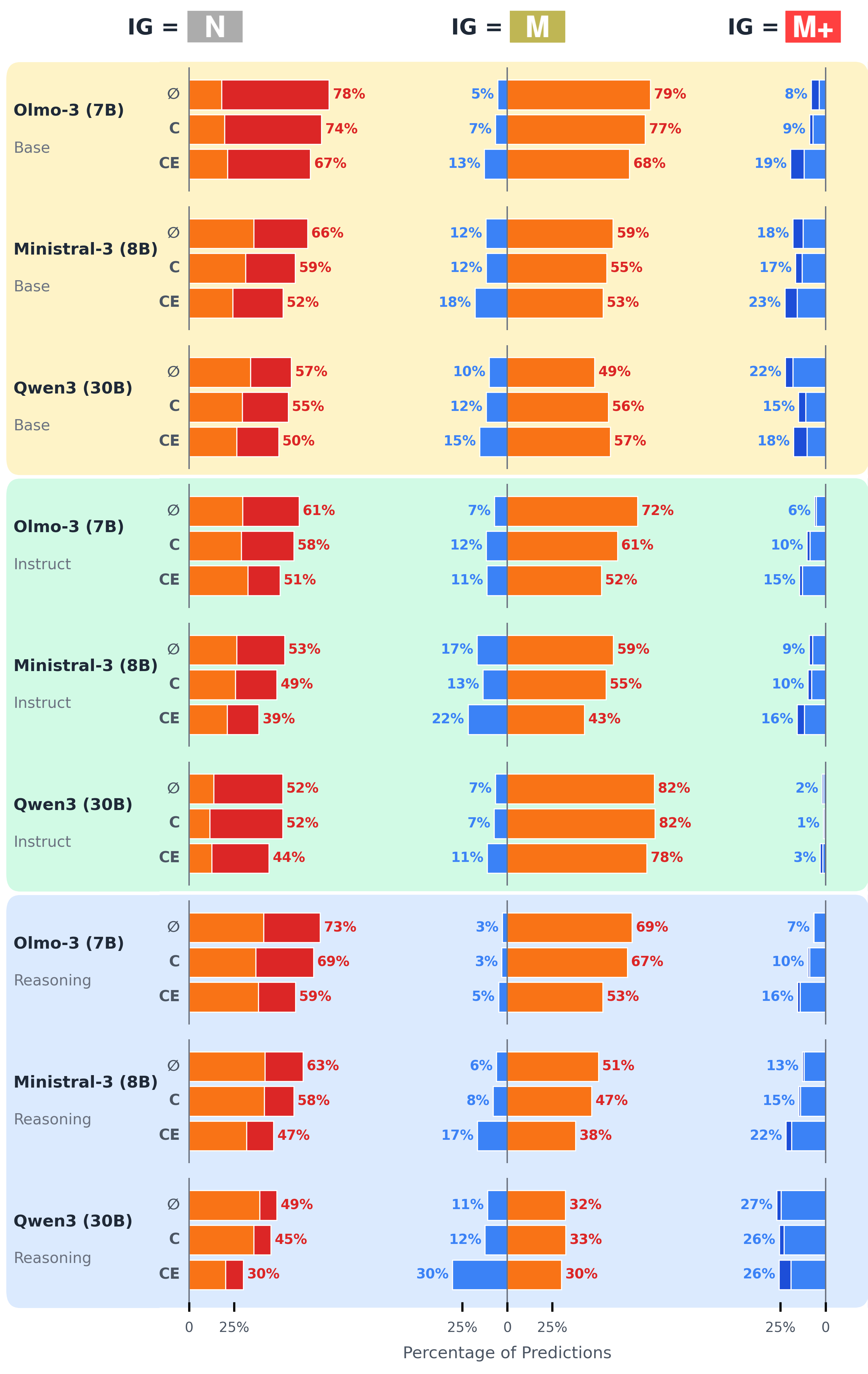}
\caption{Conditional error rates for distress severity by IG consensus. Each panel shows the rate of over-estimation (rightward bars) and under-estimation (leftward bars) for each open-weight model across the three prompting conditions. Light shades indicate an error of one; dark shades indicate an error of two. Each variant type is distinguished by a different background color.}
\label{fig:conditional_errors}
\end{figure}

Figure~\ref{fig:conditional_errors} illustrates conditional error rates by IG consensus severity across the open-weight models and prompts. Those models systematically over-estimate when the IG severity is ``None'' or ``Mild'', achieving 44\% accuracy on average, with most errors being false positives. Performance improves substantially when IG severity is ``Moderate+'', where models achieve 85\% accuracy on average.

As established in the main text, frontier models do not share a single conditional profile. GPT-5 and Gemini~2.5~Pro over-estimate about half of IG-None posts (49.8--50.5\% and 48.0\%) and still show net over-estimation on pooled none-to-mild posts. Claude Opus~4 over-estimates only 21.5\% of IG-None posts and, on Mild posts, under-estimates more often than it over-estimates (33.6\% vs.\ 12.5\%). This pattern suggests that many LLMs struggle to calibrate the lower end of the severity spectrum, precisely the range where community-specific norms diverge most, as documented in our qualitative analysis (Appendix~\ref{app:qualitative}).

\section{Community-specific directional errors}
\label{app:subreddit-bias}

\begin{figure}[!htbp]
\centering
\includegraphics[width=1\linewidth]{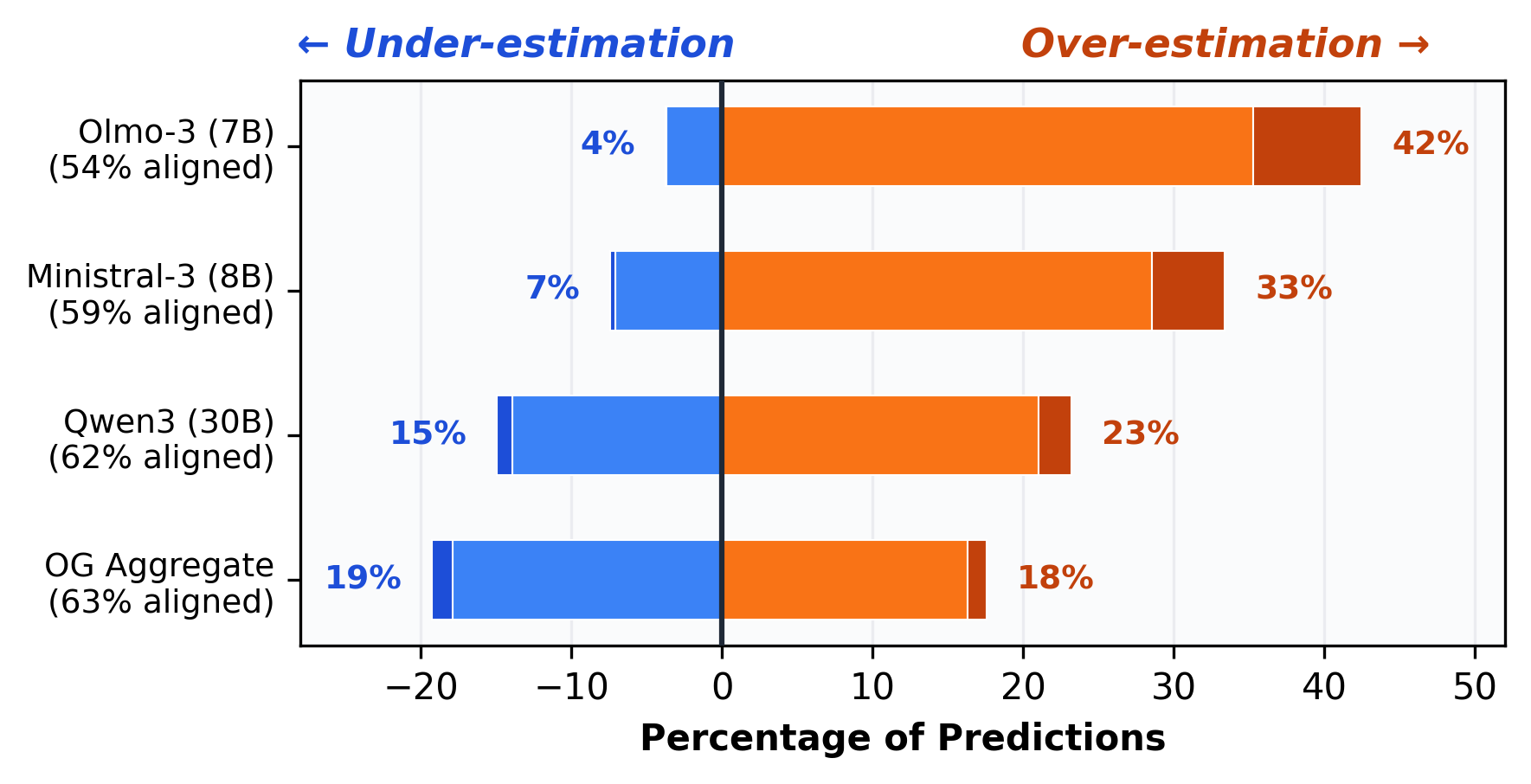}
\caption{Directional error analysis comparing under- vs. over-estimation rates for each open-weight model’s best-performing variant in the vanilla condition ($\varnothing$). Light shades indicate an error of one; dark shades indicate an error of two. Most open-weight LLMs show an over-estimation bias. Qwen3 is the least tilted of these models, but unlike the nearly symmetric OG aggregate it still over-estimates more often than it under-estimates.}
\label{fig:directional_error}
\end{figure}

\begin{figure}[!htbp]
\centering
\includegraphics[width=1\linewidth]{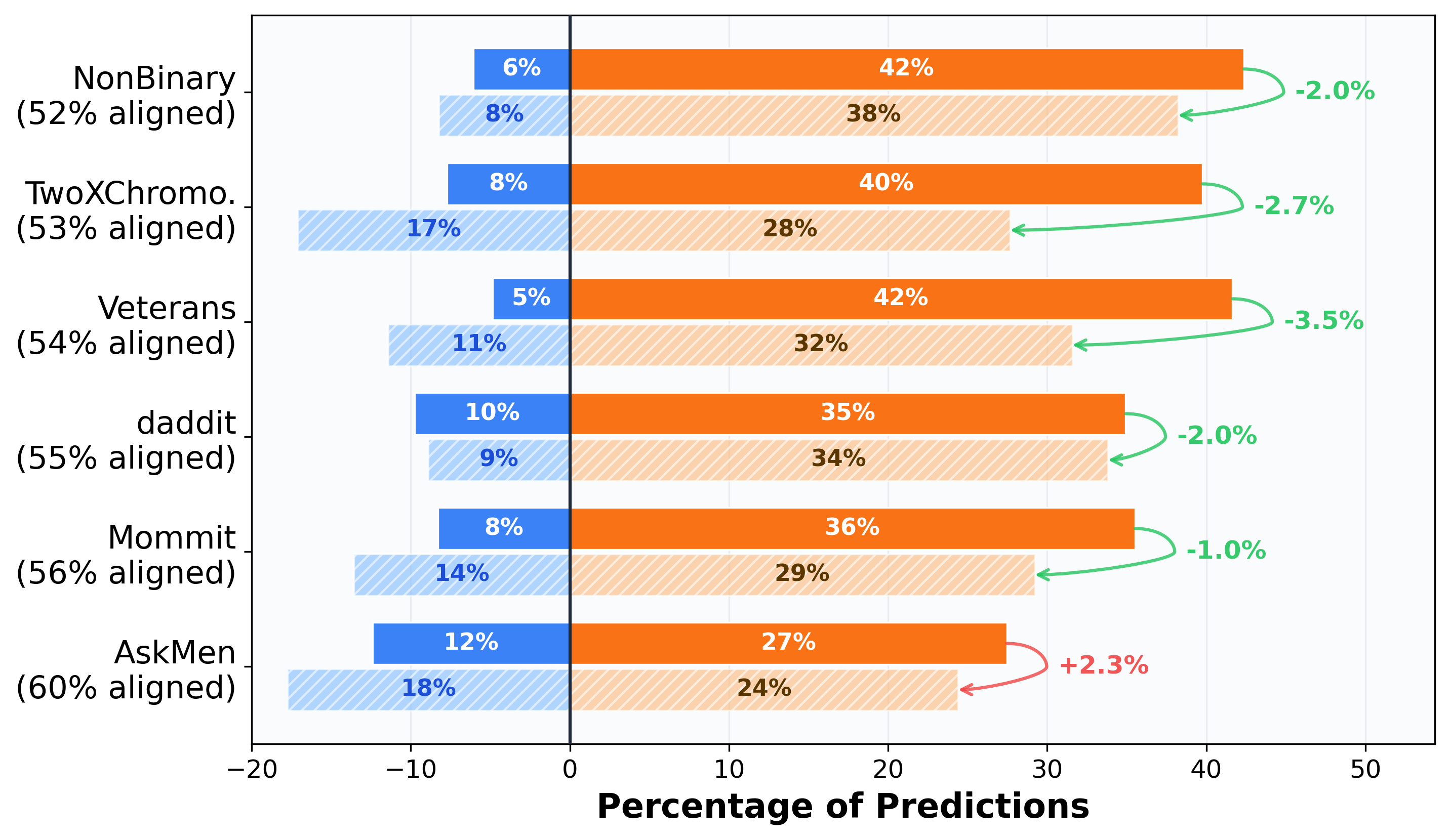}
\caption{Over- and under-estimation rates by community. Solid bars represent vanilla prompting (\(\varnothing\)), and dashed bars represent community-conditioned prompting (\(CE\)). Green arrows indicate reduced error with conditioning, while red arrows indicate increased error.}
\label{fig:subreddit_bias}
\end{figure}

Figure~\ref{fig:directional_error} presents aggregate directional error rates across models; Figure~\ref{fig:subreddit_bias} disaggregates this analysis by community, contrasting vanilla prompting with the contextualized-with-examples condition.

\section{LLM errors on disputed posts}
\label{app:disputed}
\begin{figure}[!htbp]
\centering
\includegraphics[width=1\linewidth]{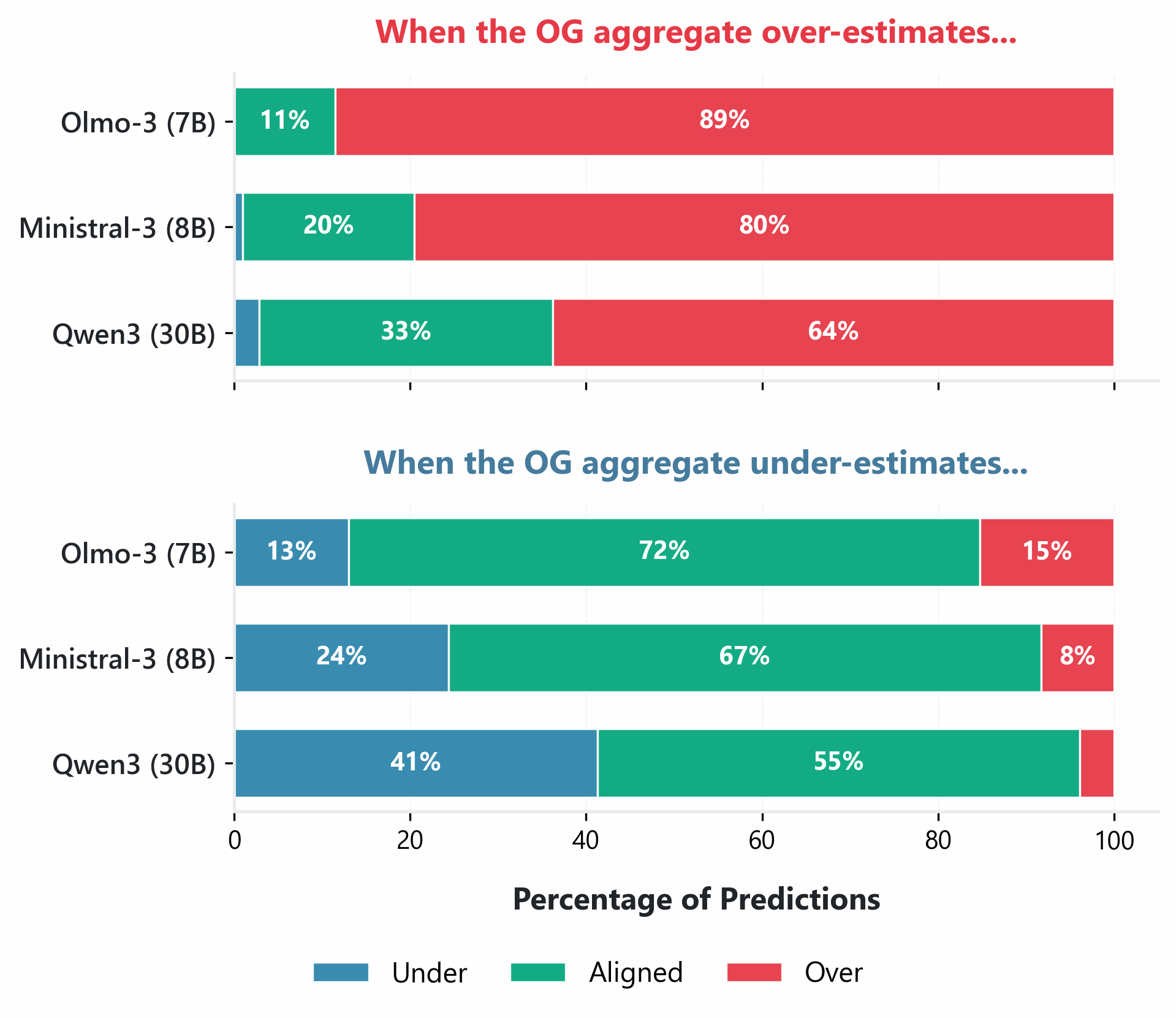}
\caption{Distribution of model predictions on disputed posts (i.e., where IG and OG aggregates disagree on severity). Bars show the percentage of predictions that under-estimate (blue), align with (green), or over-estimate (red) relative to the IG consensus.}
\label{fig:disputed}
\end{figure}

Figure~\ref{fig:disputed} outlines model behavior on posts where IG and OG aggregates diverged. On posts where the OG rated more severely than the IG ($\mathrm{n} = 210$), LLMs strongly aligned with the OG's direction: over-estimation rates ranged from 63.8\% to 88.5\%. On the other hand, on posts where the OG's aggregate labeled less severely than the IG ($\mathrm{n} = 231$), LLMs were less likely to follow the OG's direction, with under-estimation rates ranging around 13.0\% to 41.3\%. This suggests that while LLMs follow OG over-rating when it occurs, they do not similarly follow OG under-rating of more ambiguous cases.

\raggedbottom

\section{Exploratory pretraining-mixture probe using DataDecide}
\label{app:datadecide}

To probe whether miscalibration may partly originate before instruction tuning, we analyzed AI2's DataDecide Dolma1.7 1B models \cite{magnusson2025datadecide}. Holding family and scale fixed, we compared the full model against ablations removing code, math and code, Reddit, or FLAN-style data. Table~\ref{tab:datadecide} reports macro $F_1$ and per-class $F_1$.

All ablations reduced macro $F_1$ relative to the full model. The largest drops followed removal of math and code ($-0.191$), code ($-0.171$), and Reddit ($-0.129$). Class-level $F_1$ shows shifts in severity-scale use: no-code and no-math/code variants still detected some ``None'' cases but largely collapsed on ``Mild'' and failed on ``Moderate+''. No-Reddit and no-FLAN variants instead relied heavily on ``Mild'' ($F_1$ $> 0.53$) while nearly abandoning ``None'' and ``Moderate+''. These patterns suggest that pretraining mixture affects representation and calibration of the severity scale. The probe does not identify a single causal source of over-estimation, but supports a multi-stage calibration account: model-development choices before post-training matter, while base/instruct/reasoning and frontier-model results show that post-training and prompting also shape the final error profile.

\begin{table}[!htbp]
\centering
\scriptsize
\setlength{\tabcolsep}{2.2pt}
\renewcommand{\arraystretch}{1.08}

\resizebox{\columnwidth}{!}{%
\begin{tabular}{@{}lrrrrrrrr@{}}
\toprule
\textbf{Model} &
\textbf{Macro} &
\textbf{$\Delta$} &
\multicolumn{3}{c}{\textbf{Per-class $F_1$}} &
\multicolumn{3}{c}{\textbf{Pred. dist. (\%)}} \\
\cmidrule(lr){4-6} \cmidrule(lr){7-9}
& & &
\Nsev{} & \Msev{} & \Mpsev{} &
\Nsev{} & \Msev{} & \Mpsev{} \\
\midrule
Full Dolma1.7 & .316 & ---     & .392 & .445 & .111 & 39.3 & 50.6 &  9.9 \\
No Code       & .145 & $-.171$ & .380 & .056 & .000 & 95.6 &  4.4 &  0.0 \\
No Math+Code  & .125 & $-.191$ & .374 & .000 & .000 & 99.7 &  0.3 &  0.0 \\
No Reddit     & .187 & $-.129$ & .014 & .546 & .000 &  1.2 & 98.7 &  0.0 \\
No Flan       & .262 & $-.055$ & .240 & .536 & .008 & 14.4 & 84.3 &  0.2 \\
\bottomrule
\end{tabular}%
}

\caption{DataDecide pretraining-mixture probe. Macro/per-class $F_1$ and predicted-label distributions for distress severity.}
\label{tab:datadecide}
\end{table}

\section{Random-stratum LLM performance}
\label{app:random-llm}

Table~\ref{tab:random-llm-main} reports macro $F_1$ and directional error rates (including under-estimation) on the random stratum (240 posts drawn before any weak supervision, yielding 237 posts with valid predictions) alongside the full-sample values for the same vanilla configurations.

\noindent\begin{minipage}{\columnwidth}
\centering
\scriptsize
\setlength{\tabcolsep}{3pt}
\renewcommand{\arraystretch}{1.08}

\resizebox{\columnwidth}{!}{%
\begin{tabular}{@{}lrrrrrr@{}}
\toprule
& \multicolumn{3}{c}{\textbf{Full Sample}} &
\multicolumn{3}{c}{\textbf{Random Stratum}} \\
\cmidrule(lr){2-4} \cmidrule(lr){5-7}
\textbf{Model} & \textbf{Macro $F_1$} & \textbf{Over} & \textbf{Under}
& \textbf{Macro $F_1$} & \textbf{Over} & \textbf{Under} \\
\midrule
\multicolumn{7}{@{}l}{\textit{Closed-source}} \\
\quad GPT-5 (minimal)     & .615 & 18.6\% & 19.6\% & .598 & 14.3\% & 20.3\% \\
\quad GPT-5 (high)        & .595 & 18.4\% & 22.2\% & .584 & 14.3\% & 19.8\% \\
\quad Gemini 2.5 Pro      & .609 & 28.5\% &  9.2\% & .625 & 19.4\% & 14.3\% \\
\quad Claude Opus 4       & .573 &  9.6\% & 33.7\% & .531 &  6.3\% & 30.0\% \\
\midrule
\multicolumn{7}{@{}l}{\textit{Open-source (Reasoning)}} \\
\quad Qwen3-30B-A3B           & .612 & 23.2\% & 14.9\% & .619 & 18.1\% & 16.0\% \\
\quad Ministral-3-8B      & .559 & 33.4\% &  7.4\% & .622 & 24.9\% & 11.0\% \\
\quad Olmo-3-7B           & .478 & 42.5\% &  3.8\% & .556 & 38.4\% &  5.1\% \\
\midrule
\multicolumn{7}{@{}l}{\textit{Open-source (Instruct)}} \\
\quad Qwen3-30B-A3B           & .485 & 42.3\% &  3.3\% & .505 & 32.1\% &  7.2\% \\
\quad Ministral-3-8B      & .517 & 34.2\% &  9.8\% & .522 & 27.8\% & 15.2\% \\
\quad Olmo-3-7B           & .488 & 41.1\% &  5.1\% & .526 & 35.9\% &  8.4\% \\
\bottomrule
\end{tabular}%
}

\captionof{table}{LLM performance on the random stratum and full sample.}
\label{tab:random-llm-main}
\end{minipage}

\end{document}